\documentclass[letterpaper]{article} 
\usepackage{aaai2026}  
\usepackage{times}  
\usepackage{helvet}  
\usepackage{courier}  
\usepackage[hyphens]{url}  
\usepackage{graphicx} 
\usepackage{natbib}  
\usepackage{caption} 
\usepackage{algorithm}
\usepackage{algorithmic}

\usepackage{amsmath}
\usepackage{amssymb}
\usepackage{multirow}
\usepackage{booktabs}
\usepackage{bbding}
\usepackage[table,xcdraw]{xcolor}
\usepackage{newfloat}
\usepackage{listings}
\DeclareCaptionStyle{ruled}{labelfont=normalfont,labelsep=colon,strut=off} 
\floatstyle{ruled}
\newfloat{listing}{tb}{lst}{}
\floatname{listing}{Listing}
\usepackage{amssymb}
\usepackage{pifont}
\usepackage{xcolor}
\newcommand{\cmark}{\textcolor{green}{\ding{51}}}
\newcommand{\xmark}{\textcolor{red}{\ding{55}}}

\definecolor{my_green}{HTML}{F0FCEC}
\definecolor{my_blue}{HTML}{E7EDFF}

\title{SAQ-SAM: Semantically-Aligned Quantization for Segment Anything Model}
\author{
    Jing Zhang\textsuperscript{\rm 1, \rm 2}, Zhikai Li\textsuperscript{\rm 1,}\footnote{Corresponding authors.}, Chengzhi Hu\textsuperscript{\rm 1, \rm 2}, Xuewen Liu\textsuperscript{\rm 1, \rm 2}, Qingyi Gu\textsuperscript{\rm 1,}\footnotemark[1]
}
\affiliations{
    \textsuperscript{\rm 1}Institute of Automation, Chinese Academy of Sciences\\
    \textsuperscript{\rm 2}School of Artificial Intelligence, University of Chinese Academy of Sciences\\
    \{zhangjing2024, lizhikai2020, huchengzhi2024, liuxuewen2023, qingyi.gu\}@ia.ac.cn

}

\usepackage{bibentry}

\begin{document}

\maketitle

\begin{abstract}
Segment Anything Model (SAM) exhibits remarkable zero-shot segmentation capability; however, its prohibitive computational costs make edge deployment challenging. Although post-training quantization (PTQ) offers a promising compression solution, existing methods yield unsatisfactory results when applied to SAM, owing to its specialized model components and promptable workflow: 
(i) The mask decoder's attention exhibits extreme activation outliers, and we find that aggressive clipping (even 100$\times$), without smoothing or isolation, is effective in suppressing outliers while maintaining performance. Unfortunately, traditional distribution-based metrics (e.g., MSE) fail to provide such large-scale clipping. 
(ii) Existing quantization reconstruction methods neglect semantic interactivity of SAM, leading to misalignment between image feature and prompt intention.
To address the above issues, we propose SAQ-SAM in this paper, which boosts PTQ for SAM from the perspective of semantic alignment.
Specifically, we propose Perceptual-Consistency Clipping, which exploits attention focus overlap to promote aggressive clipping while preserving semantic capabilities. 
Furthermore, we propose Prompt-Aware Reconstruction, which incorporates image-prompt interactions by leveraging cross-attention in mask decoder, thus facilitating alignment in both distribution and semantic. 
Moreover, to ensure the interaction efficiency, we design a layer-skipping strategy for image tokens in encoder.
Extensive experiments are conducted on various SAM sizes and tasks, including instance segmentation, oriented object detection, and semantic segmentation, and the results show that our method consistently exhibits advantages.
For example, when quantizing SAM-B to 4-bit, SAQ-SAM achieves 11.7\% higher mAP than the baseline in instance segmentation task. Code is available at \url{https://github.com/jingjing0419/SAQ-SAM}.


\end{abstract}

\section{Introduction}

Segment Anything Model~\cite{kirillov2023segment} (SAM) shows promising applications as the base model for promptable segmentation. Benefiting from sufficient pre-training and prompt-guided fine-tuning, SAM exhibits strong zero- or few-shot generalization capability to interactively segment regions of interest to the user. However, SAM's powerful representation capability is accompanied by a large number of parameters and high computational costs, limiting its potential on resource-constrained devices~\cite{zhang2024efficientvit,chen2023slimsam}.

\begin{figure}[!t]
\centering
\includegraphics[width=1\linewidth]{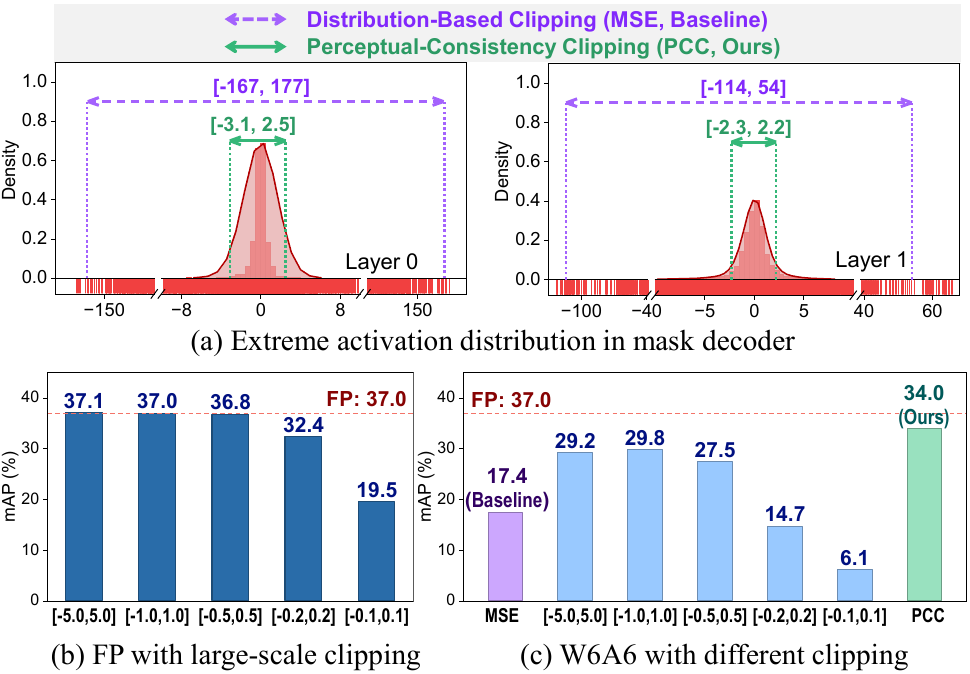}
\caption{Visualization of extreme activation distributions in mask decoder and the performance of different clipping methods. QK activations in the mask decoder show highly skewed distributions, with most data concentrated in a narrow range while outliers can exceed 180 times the normal range. MSE provides an overly wide clipping range, whereas our Perceptual-Consistency Clipping (PCC) method can identify outliers more precisely. 
}
\label{fig:distribution}
\end{figure}

\begin{figure}[!t]
\centering
\includegraphics[width=0.8\linewidth]{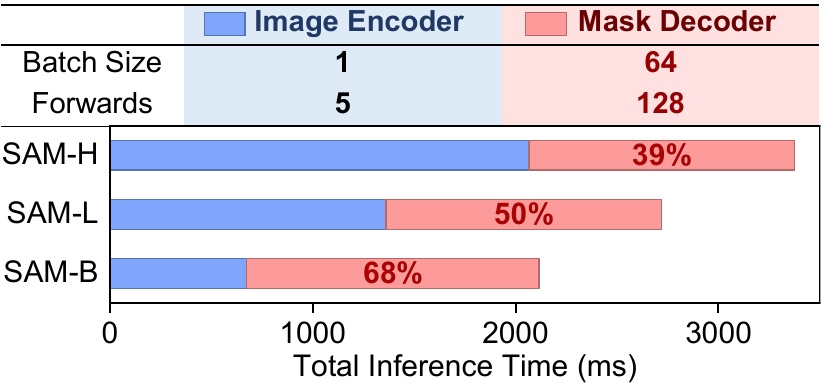}
\caption{
Comparison of total inference time between image encoder and mask decoder in semantic segmentation task. 
}
\label{fig:time_cmp}
\end{figure}

Model quantization~\cite{li2024htq,liu2024eda,xiao2024binaryvit} can reduce computational overhead and model size by converting weights and activations to low-bit integers~\cite{li2023vit,gholami2021survey,choi2018pact}, and post-training quantization (PTQ)~\cite{adaround,li2023psaq}, which efficiently calibrates quantization parameters using a small set of unlabeled data, stands out as a promising approach.
Unfortunately, SAM features the unique activation distribution and network architecture, which pose new challenges to PTQ, rendering conventional methods inadequate. 
To this end, several studies~\cite{li2024privacy} have attempted to propose specific solutions for SAM's characteristics.
For instance, PQ-SAM~\cite{pq-sam} hierarchically clusters similar channels to learn unified transformation factors. However, it is limited to the image encoder with a classical Transformer structure and lacks adaptability to the mask decoder, which features more complex connection paths.
In practice, although SAM's parameters are primarily concentrated in the image encoder, the mask decoder can also introduce significant computational overhead.
As shown in Figure \ref{fig:time_cmp}, in semantic segmentation task, given a single image, the mask decoder must perform multiple large-batch inferences for numerous prompting points, whereas the image encoder requires much fewer forward passes~\cite{chen2023semantic}.
To this end, PTQ4SAM~\cite{ptq4sam} is proposed for quantization bottleneck in the mask decoder, which merges bimodal distributions in post-key-linear activations through QK weight transformation. However, it still seeks to align full-precision (FP) model at distribution level, leading to unsatisfactory performance, especially for low-bit quantization.

In this paper, we comprehensively investigate SAM's model architecture and interactivity in promptable segmentation, and find that the semantic misalignment constitutes the main challenge for low-bit quantization. In the following, we will discuss the misalignment issues in two critical techniques of PTQ: outlier handling and reconstruction.

\begin{figure}[!t]
\centering
\includegraphics[width=0.8\linewidth]{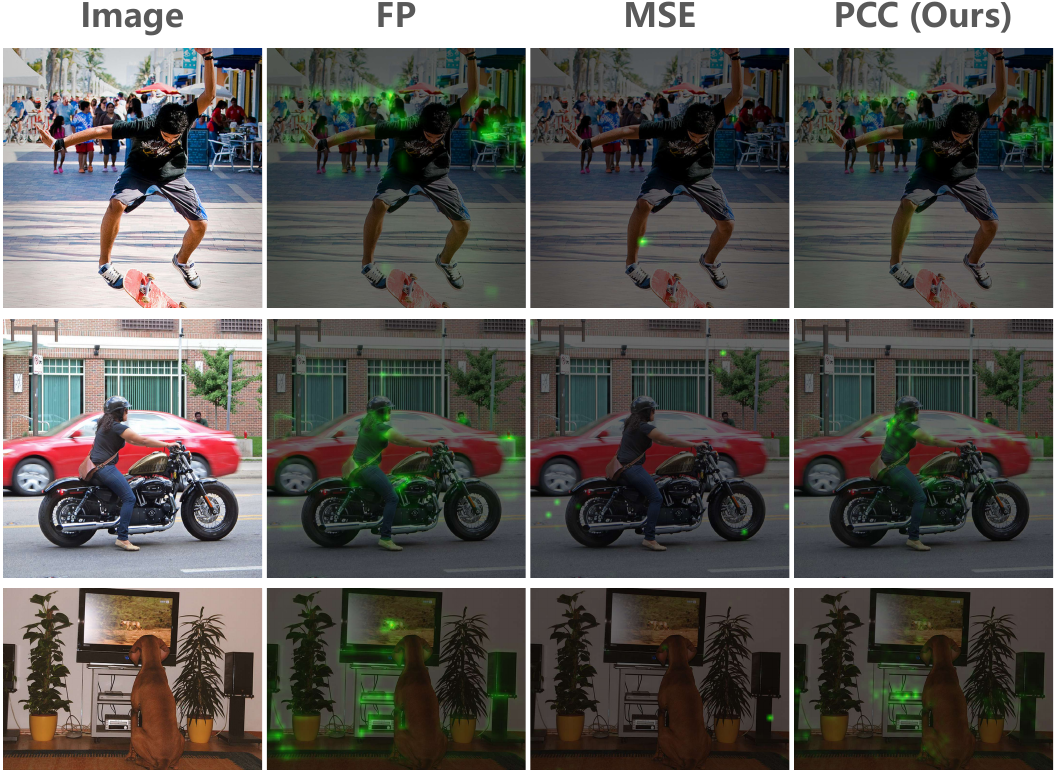}
\caption{Attention heatmaps in the mask decoder with different quantization clipping methods. The distribution-aligned MSE leads to significant attention degradation, whereas our semantic-aligned PCC maintains the consistency with the FP model.
}
\label{fig:attn_heat}
\end{figure}

\textbf{i) Semantic Misalignment in Outlier Handling.} 
QK activations in SAM's mask decoder exhibits extreme outliers, as shown in Figure \ref{fig:distribution}(a) and Appendix \ref{sec:more_distribution}. 
Fortunately, we find an interesting phenomenon: aggressively clipping these outliers has little impact on final segmentation performance. As shown in Figure \ref{fig:distribution}(b), clipping the skewed distributed activation from [-167, 177] and [-114, 54] to [-1, 1] has little impact.
More importantly, such large-scale range reduction improves quantization resolution, thereby reducing quantization errors.
To this end, we perform a grid search of large-scale clipping intervals for quantization calibration, as shown in Figure \ref{fig:distribution}(c), where [-1, 1] achieves 12.4\% performance improvement compared with traditional MSE metric.
Thus, distribution-based metrics are no longer applicable to segmentation tasks, resulting in severe semantic degradation, as shown in Figure \ref{fig:attn_heat}. 
As a result, it is crucial to establish a semantic level metric to drive optimal-scale clipping.

\textbf{ii) Semantic Misalignment in Reconstruction.} 
Quantization reconstruction is a prevalent method for restoring model performance. However, existing methods typically target at simple tasks with single branch model architecture, thus exhibiting suboptimal performance in promptable segmentation task with encoder-decoder architecture.
This is because, locally reconstructing the FP responses in the image encoder overlooks the semantic intent of prompts, thus causing feature distortions in visual-prompt interactions and ultimately compromising segmentation accuracy.
To this end, to ensure quantization accuracy, it is crucial to facilitate semantic alignment during reconstruction.

With the above insights, we propose SAQ-SAM, which significantly improves PTQ performance for SAM by promoting semantic-level alignment with the FP model. 
Specifically, for outlier suppression, we propose Perceptual-Consistency Clipping (PCC) for QK activations. By proactively measuring the quantization-induced deviation in attention focus, PCC achieves magnitude-independent outlier suppression with semantic alignment.
As illustrated in Figure \ref{fig:attn_heat}, PCC evidently mitigates the attention functional degradation issue.
Further, we propose Prompt-Aware Reconstruction (PAR), which learns quantization parameters under the guidance of prompts. By leveraging SAM’s off-the-shell mask decoder to facilitate interaction between prompts and image tokens, PAR reconstructs the interaction responses supervised by the FP model. This preserves the correspondence between visual features and prompts, thus facilitating alignment in both
distribution and semantic.
The main contributions are summarized as follows:

\begin{itemize}
\item We observe that aggressively clipping outliers in SAM has minimal impact on segmentation performance. With this insight, we propose Perceptual-Consistency Clipping, which utilizes attention focus deviation as a metric to guide large-scale clipping, and thus suppressing outliers while maintaining semantic alignment.

\item We propose Prompt-Aware Reconstruction, which integrates image-prompt interactions into reconstruction. This enables the quantization model to align with the FP model in both feature modeling and prompt following. We also design a layer-skipping strategy of image tokens to ensure interaction efficiency.

\item We conduct extensive experiments on SAMs of various sizes across three mainstream tasks, 
demonstrating the consistent superiority over baseline methods. For instance, in instance segmentation, our 4-bit SAQ-SAM achieves an average 14\% mAP improvement on SAM-B and maintains nearly lossless accuracy on SAM-L.
\end{itemize}

\begin{figure*}[!t]
\centering
\includegraphics[width=1\linewidth]{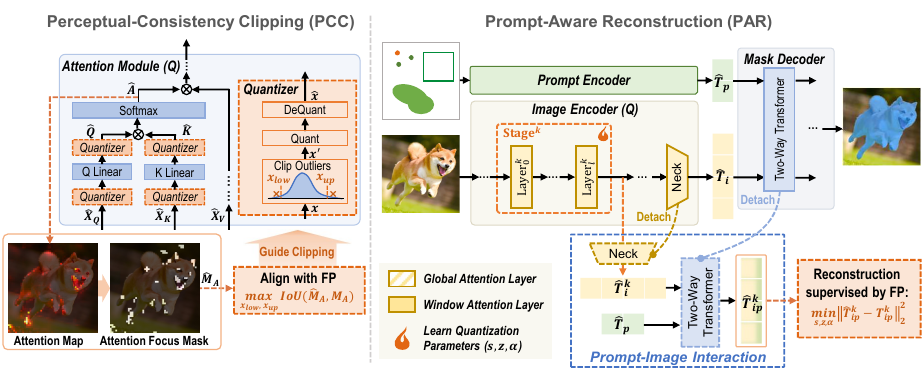}
\caption{
Overview of SAQ-SAM. The proposed PCC guides quantization clipping of QK activation by minimizing the Attention Focus deviation from FP, thereby semantically preserving the perceptual alignment. 
Our PAR incorporates image-prompt interactions into per-stage reconstruction, utilizing the off-the-shell module in the mask decoder. 
Through minimizing the interaction response error supervised by the FP model, quantization model learns correspondence between visual features and prompt intentions, thus facilitating dual alignment at both the distributional and semantic levels.
}
\label{fig:method}
\end{figure*}

\section{Related Work}
\noindent\textbf{Segment Anything.}
Segment Anything (SAM)~\cite{kirillov2023segment,ravi2024sam}  has emerged as a versatile and promptable image segmentation tool. Pre-trained on an extensive dataset, SAM demonstrates remarkable generalization across various downstream tasks, including instance segmentation~\cite{chen2024rsprompter}, semantic segmentation~\cite{chen2023semantic}, and oriented object detection~\cite{yu2023h2rbox}. Its capabilities extend to critical domains such as medical image annotation~\cite{MedSAM,sammed2d}, where it shows significant promise as a powerful diagnostic support tool. 
Thanks to its excellent generalization capabilities, SAM is also be used as an auxiliary branch for industrial anomaly detection~\cite{li2025sam_ano}.
However, SAM's powerful capabilities come with extensive memory and computational costs. Although some studies have proposed various lightweight variants to accelerate inference~\cite{fastsam,mobilesam,efficientsam}, the model parameters of these methods are still kept in floating-point type. This prevents them from benefiting from efficient low-precision integer operation units, resulting in suboptimal efficiency.

\noindent\textbf{Post-Training Quantization.}
Model quantization~\cite{li2023qft,li2022patch, liu2025cachequant} reduces memory and computation costs by converting floating-point values to integers. To mitigate the performance degradation caused by precision loss, researchers have made various efforts, among which PTQ efficiently uses a handful of unlabeled samples to set quantization parameters~\cite{lin2024awq,liu2024dilatequant,li2022dual}.

The backbone of SAM is built on the Transformer architecture, a structure for which PTQ have been extensively researched and developed. SmoothQuant~\cite{smoothquant} employs equivalent transformations to smooth the activation distribution, mitigating the impact of harmful outliers. OmniQuant~\cite{omniquant} improves the applicability of smoothing methods by learning equivalent transformation factors and clipping ranges. RepQ-ViT~\cite{repq-vit} and RepQuant~\cite{repquant} decouple quantization from inference, bridging the gap through scale reparameterization for efficient hardware-friendly quantization. In addition to these distribution-adjustment methods, some methods exploit the characteristics of the attention mechanism. For example, PTQ-ViT~\cite{ptqvit} determines quantization clipping by maintaining the consistency of the attention rank. Other methods, such as BRECQ~\cite{brecq}, QDrop~\cite{qdrop}, and PD-Quant~\cite{pd}, aim to learn optimal quantization parameters. They reconstruct the internal response of the quantization model under the supervision of the FP model.

Despite the aforementioned methods achieved remarkable performance in dealing with classical Transformer structures, SAM's distinctive activation distribution poses new challenges, leaving their performance subpar. To this end, several works try to make improvements. To tackle the channel-wise distribution imbalance in activations, PQ-SAM~\cite{pq-sam} hierarchically clusters channels with similar distributions and learns shared transformation factors for each group, thereby reducing optimization complexity. PTQ4SAM~\cite{ptq4sam} identify and integrate the bimodal distribution in post-key-linear activation in mask decoder, and implement an adaptive granularity quantizer for post-softmax activation to adapt to different types of attention module. However, these distribution-level improvements have shown unsatisfactory performance in practice.

\section{Method}

\subsection{Preliminaries}
\textbf{Transformer Layers in SAM:}
In SAM's image encoder, the Transformer layers employ window attention and global attention alternately~\cite{win_attn}, and the former divides the image into non-overlapping windows and computes self-attention within each window without shifting. In the mask decoder, a lightweight Two-Way Transformer is utilized to update both the image embedding and prompt tokens via cross-attention, facilitating the information interaction between prompt and image. Specifically, the token-to-image cross-attention utilizes prompt tokens as queries and employs image tokens as keys and values, whereas the image-to-token cross-attention implements the inverse configuration to ensure comprehensive feature interaction.

\noindent\textbf{Quantization Calibration:}
The uniform quantizer is the most commonly used and deployment-friendly quantizer, which is defined as:
\begin{equation}
    \text{Quant}:x_{q}=\text{clip}\left (\left \lfloor \frac{x}{s}\right \rceil+z, 0, 2^{b}-1\right ),
\end{equation}
\begin{equation}
    \text{DeQuant}: \hat{x}=s\left ( x_{q} - z\right )\approx x,
\end{equation}
where $x$ and $ x_{q} $ are the floating-point and quantized values, respectively. The dequantized value $ \hat{x} $ approximates $x$. $\left \lfloor \cdot \right \rceil$ is the round-to-nearest operation. clip function truncates values outside the $b$-bit range. The quantization scale $s$ and zero-point $z$ are PTQ parameters to be searched, determined by the clipping boundaries $x_{low}$ and $x_{up}$ as follows:
\begin{equation}
    s=\frac{x_{up}-x_{low}}{2^{b}-1}, z=\left \lfloor -\frac{x_{low}}{s} \right \rceil.
\end{equation}
The process of determining clipping boundaries is crucial for PTQ performance, which is also called calibration.

\subsection{Perceptual-Consistency Clipping}

\textbf{Insight.} 
With the observation that aggressively clipping the extreme outlier in QK activations, although significantly modifying the distribution, doesn't hurt the segmentation performance, we aim to break through the limitations of distribution alignment and turn to leverage the semantic nature of attention mechanisms for PTQ calibration. 
As stated in \cite{vaswani2017attention}, in Transformers architecture, the attention mechanism captures the semantic information perceived by the model. It allocates greater focus to regions of interest, thereby reinforcing the features most relevant to the task. Building on this, it's reasonable to leverage the potential relationships modeled by attention to preserve semantic-level consistency before and after quantization.

\noindent\textbf{Attention Focus Overlap Metric.} 
Inspired by above insights, we define highly attented region as the perceptual focus and maximize the overlap of the perceptual focus before and after quantization, thereby maintaining consistency in attention perception. The procedure of the metric formulation is described as follows.

Denote the input of the attention module as $X_Q\in \mathbb{R}^{N_{q}\times D}$, $X_K,X_V\in \mathbb{R}^{N_{k}\times D}$. $N_{q}$ and $N_{k}$ are the number of query tokens and key tokens, respectively. Note that, for simplicity, the batch and head dimensions are omitted. The attention score matrix is calculated as follows:
\begin{equation}
    A_{s}=X_{Q}W_{Q}\cdotp (X_{K}W_{K})^{T}/ \sqrt{d_{h}}\in \mathbb{R}^{N_{q}\times N_{k}},
\end{equation}
where $W_{Q},W_{K}\in \mathbb{R}^{D \times d_{h}}$ are weights matrix of QK linear layer, $d_{h}$ is the dimension for each head. The attention scores are normalized into attention weight matrix through the Softmax function as follows:
\begin{equation}
    A_{w}=\text{Softmax}(A_{s})={\left( {\alpha}_{1},\dots ,{\alpha}_{N_{q}} \right )}^{T}\in \mathbb{R}^{N_{q}\times N_{k}},
\end{equation}
where attention weight vector ${{\alpha}_{i}=\left ( {\alpha}_{i,1}, \dots , {\alpha}_{i,N_k} \right ) }^{T} \in \mathbb{R}^{N_{k}} $ represent the similarity of i-th query vector ${q}_{i}\in \mathbb{R}^{d_{h}}$ to all key vectors $\left \{ k_{j}\in {\mathbb{R}}^{d_{h}}, j=1, \dots , N_{k} \right \}$. A larger weight ${\alpha}_{i,j}$ indicates that ${q}_{i}$ more closely matched to ${k}_{j}$, and the model considers the corresponding value vector $v_j$ to be more important for the task at hand.

Given the above attention scores, we distinguish salient regions and define them as the \textit{\textbf{Attention Focus}}. A threshold factor $\theta \in (0,1)$ is used to filter the significant values, resulting in a binarized \textit{\textbf{Attention Focus Mask}} as follows:
\begin{equation}
    M_{A}=M(A_{w})=\textbf{1}\left\{A_{w}>\theta\cdot \max\left ( A_{w}\right ) \right\}\in\mathbb{R}^{N_{q}\times N_{k}},
\end{equation}
where $\textbf{1} \left \{  \cdot \right \}$ represents an indicator function that returns 1 if the condition is true, otherwise 0.

The Attention Focus Mask implies critical perceptual information modeled by this attention module. On this basis, we can maintain the consistency of focusing patterns with the FP model, and thus enabling alignment at the semantic level. Specifically, we define the \textbf{\textit{Attention Focus Overlap}} metric, which calculate the overlap of the Attention Focus Mask before and after quantization as follows:
\begin{equation}
    \text{IoU}_{\text{AF}}\left ( A_{w},\hat{A}_{w}\right )=\frac{\left |M_{A} \cap \hat{M}_{A}\right |}{\left |M_{A} \cup  \hat{M}_{A}\right |},
\end{equation}
where the variables with a hat superscript ($\hat{A}_{w}$/$\hat{M}_{A}$)  denote the scores/masks from the quantized model. This metric lies in the range (0,1), and the distance function for PCC is defined as:
\begin{equation}
    \text{Dist}_{pcc}=1- \text{IoU}_{\text{AF}}\left ( A_{w},\hat{A}_{w}\right ).
\end{equation}

Then, we leverage the defined Attention Focus Overlap metric to determine the optimal clipping boundaries (i.e. $x_{low}$ and $x_{up}$) for QK activations. Unlike conventional methods limited by distributional matching, our proposed clipping metric operates at a semantic level, enabling more functionally effective clipping decisions.  
As shown in Figure \ref{fig:attn_heat}, while MSE-based calibration causes severe attention degradation, our PCC achieves closer results to the FP model, efficiently preserves the semantic capabilities of the quantized attention module.

\subsection{Prompt-Aware Reconstruction}

\textbf{Insight.} 
Quantization reconstruction methods typically learn the quantization parameters $\left \{ s,z,\alpha \right \}$ by locally minimize the error of block response under the supervision of the FP model as follows:
\begin{equation}
    \min_{s,z,\alpha}{{\left \| \hat{O^{l}}-O^{l}\right \|}^2_{2}},
\end{equation}
where $O^l, {\hat{O}}^l$ are the outputs of $l$-th block from the FP model and quantized model, respectively. $\alpha$ is the adaptive rounding factor, an extra weight quantization parameter introduced by AdaRound~\cite{adaround}. 
While this method effectively enhances PTQ performance on conventional models, it struggles with the SAM model due to severe overfitting to pure image information. Specifically, Designed to follow the user's prompt, SAM's image embeddings interacting with prompts embeddings in mask decoder.
Therefore, simply considering local visual response overlooks the prompt's intent, potentially introducing redundant information that disrupts the image-prompt interaction, resulting in distorted segmentation results.

\noindent\textbf{Interaction Response Reconstruction.} To address the aforementioned issue, we incorporate the interaction between prompt and image tokens into the reconstruction process, enabling effective image-prompt matching under the supervision of the FP model.
Specifically, instead of aligning the raw visual response, we reconstruct the hybrid image tokens that integrate the prompt information by interacting in the cross-attention of mask decoder.

As illustrated in Figure \ref{fig:method}, we utilize the off-the-shelf cross attention module in SAM's mask decoder to incorporate prompt information for the image tokens, without additional components or extra training.
In this way, the hybrid image tokens can be obtained as follows: 
\begin{equation}
T_{ip}^{k}=\text{TwoWayTransformer}(T_{i}^{k},T_{p}),
\end{equation}
where $T_{p}$ denotes prompt tokens encoded by prompt encoder, $T_{i}^{k}$ denotes image tokens derived from the outputs of the $k$-th stage. TwoWayTransformer refers to the Two-Way Transformer module detached from the mask decoder, which is designed for efficient image-prompt interaction. Subsequently, these tokens are reconstructed to learn the quantization parameters, i.e., minimizing the L2 distance between the hybrid image tokens and the corresponding FP response as follows:
\begin{equation}
\min_{s,z,\alpha}{{\left \| {\hat{T}}_{ip}^{k}-T_{ip}^{k}\right \|}^2_{2}},
\end{equation}
where $T_{ip}^{k}$ and ${\hat{T}}_{ip}^{k}$ are tokens from the FP model and the quantized model, respectively.

\noindent\textbf{Layer Skipping Interaction.}
To boost the efficiency of reconstruction, we customize two designs based on SAM's characteristics. 
First, we partition the Transformer layers into multiple stages and adopt stage-wise learning, where the quantization parameters within each stage are jointly optimized. Previous work~\cite{interactions-across-block} has demonstrated the superiority of jointly optimizing multiple quantization blocks in reconstruction, as it effectively captures weight correlations across blocks. Building on this insight and SAM's architecture, we define global attention layers as boundaries for stage partitioning. For example, in SAM-B, layers $L2, L5, L8$ and $L11$ employ global attention, while the remaining layers utilize window attention. Accordingly, layers $\{L0, L1, L2\}$ constitute stage 0, and this pattern repeats for subsequent stages.

Furthermore, to promote efficient image-prompt interactions, the outputs of learning stage skip subsequent layers and directly pass the neck for interaction:
\begin{equation}
    T_{i}^{k}=\text{Neck}((\textstyle\prod_{i=0}^{k}\text{Stage}^{k})(E_{i})),
\end{equation}
where $E_{i}$ denotes patch embeddings, $\text{Stage}^{k}$ denotes the k-th stage, $\textstyle\prod_{i=0}^{k}\text{Stage}^{k}$ represents the sequential composition of stages from stage 0 to stage k, $\text{Neck}$ denotes SAM's Neck module for dimension reduction of image features.
This layer-skipping strategy avoids prohibitively high computational cost of complete forward passes and gradient backpropagation, thus introducing slight additional overhead compared to traditional local response reconstruction. In addition to efficiency benefits, skipping deeper layers also brings performance advantages. With shortened optimization paths, it not only preserves local distribution properties, but also avoids the potential loss of interactive information due to long propagation processes.

\begin{figure}[t]
\centering
\includegraphics[width=0.8\linewidth]{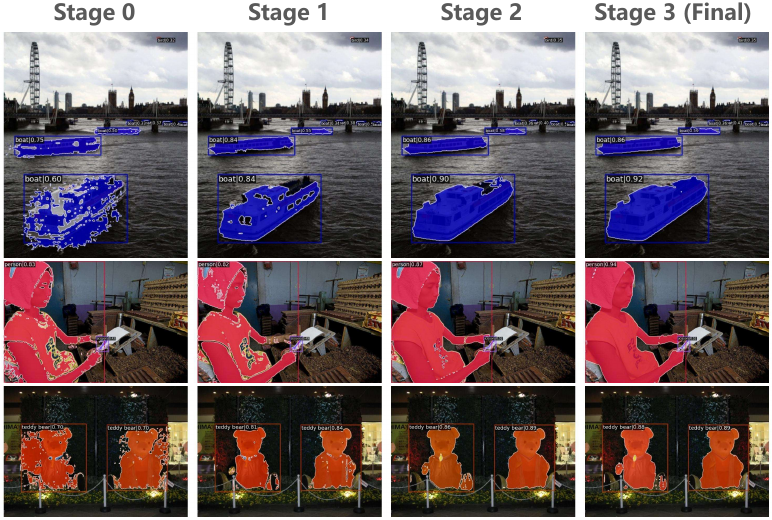}
\caption{
    Segmentation results with image tokens from different stages. The output features of each stage are capable of skipping subsequent propagate while ensuring competent segmentation, with quality improving at deeper stages.
}
\label{fig:stage_out}
\end{figure}

To validate the feasibility of using these layer-skipping tokens directly for interaction, we tentatively use them as the final visual representation for mask prediction. The results shown in Figure \ref{fig:stage_out} indicate that these immature tokens can produce rational segmentation results without further processing by deeper layers. This provides implicit evidence that the layer-skipping design is capable of effective interaction, with the outputs of each stage functioning as image embeddings at different levels of semantic granularity.


\begin{table*}[!ht]\scriptsize
\centering
\caption{Quantization results of instance segmentation on COCO dataset with various detectors providing prompts. We abbreviate `Precision' as `Prec.', `W/A'=$x/y$ denotes the weights and activations are quantized to $x$ bit and $y$ bit, respectively, and `-' indicates the mAP is below 1. 
Rows with \colorbox{my_green}{green background} indicate statistic-based methods, while \colorbox{my_blue}{blue background} indicate learning-based methods.
The bolded values represent the best-performing results under the same settings.}
\setlength{\tabcolsep}{3.4pt}
\begin{tabular}{@{}c|cccccc|cccccc|cccccc|cccccc@{}}
\toprule
Detector & \multicolumn{6}{c|}{Faster R-CNN} & \multicolumn{6}{c|}{YOLOX} & \multicolumn{6}{c|}{H-Deformable-DETR} & \multicolumn{6}{c}{DINO} \\ \midrule
Model & \multicolumn{2}{c|}{SAM-B} & \multicolumn{2}{c|}{SAM-L} & \multicolumn{2}{c|}{SAM-H} & \multicolumn{2}{c|}{SAM-B} & \multicolumn{2}{c|}{SAM-L} & \multicolumn{2}{c|}{SAM-H} & \multicolumn{2}{c|}{SAM-B} & \multicolumn{2}{c|}{SAM-L} & \multicolumn{2}{c|}{SAM-H} & \multicolumn{2}{c|}{SAM-B} & \multicolumn{2}{c|}{SAM-L} & \multicolumn{2}{c}{SAM-H} \\ \midrule
\rowcolor[HTML]{EFEFEF} 
FP & \multicolumn{2}{c|}{\cellcolor[HTML]{EFEFEF}33.4} & \multicolumn{2}{c|}{\cellcolor[HTML]{EFEFEF}36.4} & \multicolumn{2}{c|}{\cellcolor[HTML]{EFEFEF}37.2} & \multicolumn{2}{c|}{\cellcolor[HTML]{EFEFEF}37.0} & \multicolumn{2}{c|}{\cellcolor[HTML]{EFEFEF}40.0} & \multicolumn{2}{c|}{\cellcolor[HTML]{EFEFEF}41.0} & \multicolumn{2}{c|}{\cellcolor[HTML]{EFEFEF}38.2} & \multicolumn{2}{c|}{\cellcolor[HTML]{EFEFEF}41.5} & \multicolumn{2}{c|}{\cellcolor[HTML]{EFEFEF}42.0} & \multicolumn{2}{c|}{\cellcolor[HTML]{EFEFEF}44.5} & \multicolumn{2}{c|}{\cellcolor[HTML]{EFEFEF}48.6} & \multicolumn{2}{c}{\cellcolor[HTML]{EFEFEF}49.1} \\ \midrule
Prec. (W/A) & \multicolumn{1}{c|}{6/6} & \multicolumn{1}{c|}{4/4} & \multicolumn{1}{c|}{6/6} & \multicolumn{1}{c|}{4/4} & \multicolumn{1}{c|}{6/6} & 4/4 & \multicolumn{1}{c|}{6/6} & \multicolumn{1}{c|}{4/4} & \multicolumn{1}{c|}{6/6} & \multicolumn{1}{c|}{4/4} & \multicolumn{1}{c|}{6/6} & 4/4 & \multicolumn{1}{c|}{6/6} & \multicolumn{1}{c|}{4/4} & \multicolumn{1}{c|}{6/6} & \multicolumn{1}{c|}{4/4} & \multicolumn{1}{c|}{6/6} & 4/4 & \multicolumn{1}{c|}{6/6} & \multicolumn{1}{c|}{4/4} & \multicolumn{1}{c|}{6/6} & \multicolumn{1}{c|}{4/4} & \multicolumn{1}{c|}{6/6} & 4/4 \\ \midrule
\rowcolor[HTML]{F0FCEC} 
MinMax & \multicolumn{1}{c|}{\cellcolor[HTML]{F0FCEC}9.2} & \multicolumn{1}{c|}{\cellcolor[HTML]{F0FCEC}-} & \multicolumn{1}{c|}{\cellcolor[HTML]{F0FCEC}32.9} & \multicolumn{1}{c|}{\cellcolor[HTML]{F0FCEC}-} & \multicolumn{1}{c|}{\cellcolor[HTML]{F0FCEC}31.9} & - & \multicolumn{1}{c|}{\cellcolor[HTML]{F0FCEC}10.7} & \multicolumn{1}{c|}{\cellcolor[HTML]{F0FCEC}-} & \multicolumn{1}{c|}{\cellcolor[HTML]{F0FCEC}37.5} & \multicolumn{1}{c|}{\cellcolor[HTML]{F0FCEC}-} & \multicolumn{1}{c|}{\cellcolor[HTML]{F0FCEC}36.1} & - & \multicolumn{1}{c|}{\cellcolor[HTML]{F0FCEC}10.9} & \multicolumn{1}{c|}{\cellcolor[HTML]{F0FCEC}-} & \multicolumn{1}{c|}{\cellcolor[HTML]{F0FCEC}38.6} & \multicolumn{1}{c|}{\cellcolor[HTML]{F0FCEC}-} & \multicolumn{1}{c|}{\cellcolor[HTML]{F0FCEC}37.3} & - & \multicolumn{1}{c|}{\cellcolor[HTML]{F0FCEC}11.2} & \multicolumn{1}{c|}{\cellcolor[HTML]{F0FCEC}-} & \multicolumn{1}{c|}{\cellcolor[HTML]{F0FCEC}44.7} & \multicolumn{1}{c|}{\cellcolor[HTML]{F0FCEC}-} & \multicolumn{1}{c|}{\cellcolor[HTML]{F0FCEC}42.8} & - \\
\rowcolor[HTML]{F0FCEC} 
Percentile & \multicolumn{1}{c|}{\cellcolor[HTML]{F0FCEC}10.9} & \multicolumn{1}{c|}{\cellcolor[HTML]{F0FCEC}-} & \multicolumn{1}{c|}{\cellcolor[HTML]{F0FCEC}33.5} & \multicolumn{1}{c|}{\cellcolor[HTML]{F0FCEC}-} & \multicolumn{1}{c|}{\cellcolor[HTML]{F0FCEC}32.0} & - & \multicolumn{1}{c|}{\cellcolor[HTML]{F0FCEC}12.0} & \multicolumn{1}{c|}{\cellcolor[HTML]{F0FCEC}-} & \multicolumn{1}{c|}{\cellcolor[HTML]{F0FCEC}38.0} & \multicolumn{1}{c|}{\cellcolor[HTML]{F0FCEC}-} & \multicolumn{1}{c|}{\cellcolor[HTML]{F0FCEC}36.3} & - & \multicolumn{1}{c|}{\cellcolor[HTML]{F0FCEC}12.3} & \multicolumn{1}{c|}{\cellcolor[HTML]{F0FCEC}-} & \multicolumn{1}{c|}{\cellcolor[HTML]{F0FCEC}39.0} & \multicolumn{1}{c|}{\cellcolor[HTML]{F0FCEC}-} & \multicolumn{1}{c|}{\cellcolor[HTML]{F0FCEC}37.5} & - & \multicolumn{1}{c|}{\cellcolor[HTML]{F0FCEC}14.0} & \multicolumn{1}{c|}{\cellcolor[HTML]{F0FCEC}-} & \multicolumn{1}{c|}{\cellcolor[HTML]{F0FCEC}45.4} & \multicolumn{1}{c|}{\cellcolor[HTML]{F0FCEC}-} & \multicolumn{1}{c|}{\cellcolor[HTML]{F0FCEC}43.1} & - \\
\rowcolor[HTML]{F0FCEC} 
OMSE & \multicolumn{1}{c|}{\cellcolor[HTML]{F0FCEC}11.9} & \multicolumn{1}{c|}{\cellcolor[HTML]{F0FCEC}-} & \multicolumn{1}{c|}{\cellcolor[HTML]{F0FCEC}33.9} & \multicolumn{1}{c|}{\cellcolor[HTML]{F0FCEC}5.4} & \multicolumn{1}{c|}{\cellcolor[HTML]{F0FCEC}33.1} & 7.4 & \multicolumn{1}{c|}{\cellcolor[HTML]{F0FCEC}13.5} & \multicolumn{1}{c|}{\cellcolor[HTML]{F0FCEC}-} & \multicolumn{1}{c|}{\cellcolor[HTML]{F0FCEC}38.4} & \multicolumn{1}{c|}{\cellcolor[HTML]{F0FCEC}6.1} & \multicolumn{1}{c|}{\cellcolor[HTML]{F0FCEC}37.5} & 7.8 & \multicolumn{1}{c|}{\cellcolor[HTML]{F0FCEC}15.0} & \multicolumn{1}{c|}{\cellcolor[HTML]{F0FCEC}-} & \multicolumn{1}{c|}{\cellcolor[HTML]{F0FCEC}39.6} & \multicolumn{1}{c|}{\cellcolor[HTML]{F0FCEC}6.2} & \multicolumn{1}{c|}{\cellcolor[HTML]{F0FCEC}38.6} & 7.7 & \multicolumn{1}{c|}{\cellcolor[HTML]{F0FCEC}16.6} & \multicolumn{1}{c|}{\cellcolor[HTML]{F0FCEC}-} & \multicolumn{1}{c|}{\cellcolor[HTML]{F0FCEC}45.9} & \multicolumn{1}{c|}{\cellcolor[HTML]{F0FCEC}6.8} & \multicolumn{1}{c|}{\cellcolor[HTML]{F0FCEC}44.5} & 8.3 \\
\rowcolor[HTML]{F0FCEC} 
PTQ4SAM-S & \multicolumn{1}{c|}{\cellcolor[HTML]{F0FCEC}15.4} & \multicolumn{1}{c|}{\cellcolor[HTML]{F0FCEC}-} & \multicolumn{1}{c|}{\cellcolor[HTML]{F0FCEC}35.7} & \multicolumn{1}{c|}{\cellcolor[HTML]{F0FCEC}18.1} & \multicolumn{1}{c|}{\cellcolor[HTML]{F0FCEC}36.0} & 24.1 & \multicolumn{1}{c|}{\cellcolor[HTML]{F0FCEC}17.4} & \multicolumn{1}{c|}{\cellcolor[HTML]{F0FCEC}-} & \multicolumn{1}{c|}{\cellcolor[HTML]{F0FCEC}\textbf{40.0}} & \multicolumn{1}{c|}{\cellcolor[HTML]{F0FCEC}20.6} & \multicolumn{1}{c|}{\cellcolor[HTML]{F0FCEC}\textbf{40.3}} & 26.7 & \multicolumn{1}{c|}{\cellcolor[HTML]{F0FCEC}17.9} & \multicolumn{1}{c|}{\cellcolor[HTML]{F0FCEC}-} & \multicolumn{1}{c|}{\cellcolor[HTML]{F0FCEC}41.0} & \multicolumn{1}{c|}{\cellcolor[HTML]{F0FCEC}20.9} & \multicolumn{1}{c|}{\cellcolor[HTML]{F0FCEC}\textbf{41.3}} & 27.3 & \multicolumn{1}{c|}{\cellcolor[HTML]{F0FCEC}20.4} & \multicolumn{1}{c|}{\cellcolor[HTML]{F0FCEC}-} & \multicolumn{1}{c|}{\cellcolor[HTML]{F0FCEC}47.7} & \multicolumn{1}{c|}{\cellcolor[HTML]{F0FCEC}23.1} & \multicolumn{1}{c|}{\cellcolor[HTML]{F0FCEC}48.1} & 30.5 \\
\rowcolor[HTML]{DCF5D3} 
SAQ-SAM$^{\star}$ & \multicolumn{1}{c|}{\cellcolor[HTML]{DCF5D3}\textbf{29.8}} & \multicolumn{1}{c|}{\cellcolor[HTML]{DCF5D3}\textbf{2.8}} & \multicolumn{1}{c|}{\cellcolor[HTML]{DCF5D3}\textbf{35.7}} & \multicolumn{1}{c|}{\cellcolor[HTML]{DCF5D3}\textbf{20.9}} & \multicolumn{1}{c|}{\cellcolor[HTML]{DCF5D3}\textbf{36.1}} & \textbf{25.0} & \multicolumn{1}{c|}{\cellcolor[HTML]{DCF5D3}\textbf{34.0}} & \multicolumn{1}{c|}{\cellcolor[HTML]{DCF5D3}\textbf{2.1}} & \multicolumn{1}{c|}{\cellcolor[HTML]{DCF5D3}\textbf{40.0}} & \multicolumn{1}{c|}{\cellcolor[HTML]{DCF5D3}\textbf{24.2}} & \multicolumn{1}{c|}{\cellcolor[HTML]{DCF5D3}40.2} & \textbf{28.4} & \multicolumn{1}{c|}{\cellcolor[HTML]{DCF5D3}\textbf{36.4}} & \multicolumn{1}{c|}{\cellcolor[HTML]{DCF5D3}\textbf{2.6}} & \multicolumn{1}{c|}{\cellcolor[HTML]{DCF5D3}\textbf{41.1}} & \multicolumn{1}{c|}{\cellcolor[HTML]{DCF5D3}\textbf{24.3}} & \multicolumn{1}{c|}{\cellcolor[HTML]{DCF5D3}\textbf{41.3}} & \textbf{28.2} & \multicolumn{1}{c|}{\cellcolor[HTML]{DCF5D3}\textbf{39.4}} & \multicolumn{1}{c|}{\cellcolor[HTML]{DCF5D3}\textbf{3.5}} & \multicolumn{1}{c|}{\cellcolor[HTML]{DCF5D3}\textbf{48.0}} & \multicolumn{1}{c|}{\cellcolor[HTML]{DCF5D3}\textbf{27.8}} & \multicolumn{1}{c|}{\cellcolor[HTML]{DCF5D3}\textbf{48.2}} & \textbf{31.6} \\ \midrule
\rowcolor[HTML]{E7EDFF} 
AdaRound & \multicolumn{1}{c|}{\cellcolor[HTML]{E7EDFF}23.1} & \multicolumn{1}{c|}{\cellcolor[HTML]{E7EDFF}-} & \multicolumn{1}{c|}{\cellcolor[HTML]{E7EDFF}34.3} & \multicolumn{1}{c|}{\cellcolor[HTML]{E7EDFF}8.7} & \multicolumn{1}{c|}{\cellcolor[HTML]{E7EDFF}33.7} & 14.5 & \multicolumn{1}{c|}{\cellcolor[HTML]{E7EDFF}26.4} & \multicolumn{1}{c|}{\cellcolor[HTML]{E7EDFF}-} & \multicolumn{1}{c|}{\cellcolor[HTML]{E7EDFF}38.9} & \multicolumn{1}{c|}{\cellcolor[HTML]{E7EDFF}11.1} & \multicolumn{1}{c|}{\cellcolor[HTML]{E7EDFF}38.3} & 16.7 & \multicolumn{1}{c|}{\cellcolor[HTML]{E7EDFF}27.2} & \multicolumn{1}{c|}{\cellcolor[HTML]{E7EDFF}-} & \multicolumn{1}{c|}{\cellcolor[HTML]{E7EDFF}39.9} & \multicolumn{1}{c|}{\cellcolor[HTML]{E7EDFF}8.0} & \multicolumn{1}{c|}{\cellcolor[HTML]{E7EDFF}39.4} & 16.3 & \multicolumn{1}{c|}{\cellcolor[HTML]{E7EDFF}31.2} & \multicolumn{1}{c|}{\cellcolor[HTML]{E7EDFF}1.2} & \multicolumn{1}{c|}{\cellcolor[HTML]{E7EDFF}46.6} & \multicolumn{1}{c|}{\cellcolor[HTML]{E7EDFF}8.8} & \multicolumn{1}{c|}{\cellcolor[HTML]{E7EDFF}46.0} & 18.2 \\
\rowcolor[HTML]{E7EDFF} 
BRECQ & \multicolumn{1}{c|}{\cellcolor[HTML]{E7EDFF}24.1} & \multicolumn{1}{c|}{\cellcolor[HTML]{E7EDFF}-} & \multicolumn{1}{c|}{\cellcolor[HTML]{E7EDFF}34.2} & \multicolumn{1}{c|}{\cellcolor[HTML]{E7EDFF}10.7} & \multicolumn{1}{c|}{\cellcolor[HTML]{E7EDFF}33.7} & 15.1 & \multicolumn{1}{c|}{\cellcolor[HTML]{E7EDFF}26.1} & \multicolumn{1}{c|}{\cellcolor[HTML]{E7EDFF}-} & \multicolumn{1}{c|}{\cellcolor[HTML]{E7EDFF}38.9} & \multicolumn{1}{c|}{\cellcolor[HTML]{E7EDFF}12.0} & \multicolumn{1}{c|}{\cellcolor[HTML]{E7EDFF}38.3} & 16.3 & \multicolumn{1}{c|}{\cellcolor[HTML]{E7EDFF}27.9} & \multicolumn{1}{c|}{\cellcolor[HTML]{E7EDFF}-} & \multicolumn{1}{c|}{\cellcolor[HTML]{E7EDFF}39.9} & \multicolumn{1}{c|}{\cellcolor[HTML]{E7EDFF}11.1} & \multicolumn{1}{c|}{\cellcolor[HTML]{E7EDFF}39.5} & 15.5 & \multicolumn{1}{c|}{\cellcolor[HTML]{E7EDFF}31.8} & \multicolumn{1}{c|}{\cellcolor[HTML]{E7EDFF}3.6} & \multicolumn{1}{c|}{\cellcolor[HTML]{E7EDFF}46.6} & \multicolumn{1}{c|}{\cellcolor[HTML]{E7EDFF}12.3} & \multicolumn{1}{c|}{\cellcolor[HTML]{E7EDFF}46.0} & 17.6 \\
\rowcolor[HTML]{E7EDFF} 
QDrop & \multicolumn{1}{c|}{\cellcolor[HTML]{E7EDFF}29.3} & \multicolumn{1}{c|}{\cellcolor[HTML]{E7EDFF}13.0} & \multicolumn{1}{c|}{\cellcolor[HTML]{E7EDFF}35.2} & \multicolumn{1}{c|}{\cellcolor[HTML]{E7EDFF}22.6} & \multicolumn{1}{c|}{\cellcolor[HTML]{E7EDFF}36.3} & 32.3 & \multicolumn{1}{c|}{\cellcolor[HTML]{E7EDFF}33.6} & \multicolumn{1}{c|}{\cellcolor[HTML]{E7EDFF}13.3} & \multicolumn{1}{c|}{\cellcolor[HTML]{E7EDFF}39.7} & \multicolumn{1}{c|}{\cellcolor[HTML]{E7EDFF}25.3} & \multicolumn{1}{c|}{\cellcolor[HTML]{E7EDFF}40.4} & 35.8 & \multicolumn{1}{c|}{\cellcolor[HTML]{E7EDFF}34.3} & \multicolumn{1}{c|}{\cellcolor[HTML]{E7EDFF}13.2} & \multicolumn{1}{c|}{\cellcolor[HTML]{E7EDFF}40.5} & \multicolumn{1}{c|}{\cellcolor[HTML]{E7EDFF}25.8} & \multicolumn{1}{c|}{\cellcolor[HTML]{E7EDFF}41.4} & 36.5 & \multicolumn{1}{c|}{\cellcolor[HTML]{E7EDFF}38.9} & \multicolumn{1}{c|}{\cellcolor[HTML]{E7EDFF}11.2} & \multicolumn{1}{c|}{\cellcolor[HTML]{E7EDFF}47.5} & \multicolumn{1}{c|}{\cellcolor[HTML]{E7EDFF}27.5} & \multicolumn{1}{c|}{\cellcolor[HTML]{E7EDFF}48.3} & 41.7 \\
\rowcolor[HTML]{E7EDFF} 
PTQ4SAM-L & \multicolumn{1}{c|}{\cellcolor[HTML]{E7EDFF}30.3} & \multicolumn{1}{c|}{\cellcolor[HTML]{E7EDFF}16.0} & \multicolumn{1}{c|}{\cellcolor[HTML]{E7EDFF}35.8} & \multicolumn{1}{c|}{\cellcolor[HTML]{E7EDFF}28.7} & \multicolumn{1}{c|}{\cellcolor[HTML]{E7EDFF}36.5} & 33.5 & \multicolumn{1}{c|}{\cellcolor[HTML]{E7EDFF}34.3} & \multicolumn{1}{c|}{\cellcolor[HTML]{E7EDFF}18.4} & \multicolumn{1}{c|}{\cellcolor[HTML]{E7EDFF}40.3} & \multicolumn{1}{c|}{\cellcolor[HTML]{E7EDFF}31.6} & \multicolumn{1}{c|}{\cellcolor[HTML]{E7EDFF}40.7} & 37.6 & \multicolumn{1}{c|}{\cellcolor[HTML]{E7EDFF}35.1} & \multicolumn{1}{c|}{\cellcolor[HTML]{E7EDFF}17.3} & \multicolumn{1}{c|}{\cellcolor[HTML]{E7EDFF}41.2} & \multicolumn{1}{c|}{\cellcolor[HTML]{E7EDFF}32.1} & \multicolumn{1}{c|}{\cellcolor[HTML]{E7EDFF}41.6} & 38.4 & \multicolumn{1}{c|}{\cellcolor[HTML]{E7EDFF}40.4} & \multicolumn{1}{c|}{\cellcolor[HTML]{E7EDFF}14.4} & \multicolumn{1}{c|}{\cellcolor[HTML]{E7EDFF}\textbf{48.3}} & \multicolumn{1}{c|}{\cellcolor[HTML]{E7EDFF}36.6} & \multicolumn{1}{c|}{\cellcolor[HTML]{E7EDFF}48.7} & 43.9 \\
\rowcolor[HTML]{CBD8FF} 
SAQ-SAM & \multicolumn{1}{c|}{\cellcolor[HTML]{CBD8FF}\textbf{32.0}} & \multicolumn{1}{c|}{\cellcolor[HTML]{CBD8FF}\textbf{27.7}} & \multicolumn{1}{c|}{\cellcolor[HTML]{CBD8FF}\textbf{35.9}} & \multicolumn{1}{c|}{\cellcolor[HTML]{CBD8FF}\textbf{34.5}} & \multicolumn{1}{c|}{\cellcolor[HTML]{CBD8FF}\textbf{36.6}} & \textbf{35.9} & \multicolumn{1}{c|}{\cellcolor[HTML]{CBD8FF}\textbf{35.9}} & \multicolumn{1}{c|}{\cellcolor[HTML]{CBD8FF}\textbf{30.3}} & \multicolumn{1}{c|}{\cellcolor[HTML]{CBD8FF}\textbf{40.3}} & \multicolumn{1}{c|}{\cellcolor[HTML]{CBD8FF}\textbf{39.0}} & \multicolumn{1}{c|}{\cellcolor[HTML]{CBD8FF}\textbf{40.9}} & \textbf{39.9} & \multicolumn{1}{c|}{\cellcolor[HTML]{CBD8FF}\textbf{37.1}} & \multicolumn{1}{c|}{\cellcolor[HTML]{CBD8FF}\textbf{31.8}} & \multicolumn{1}{c|}{\cellcolor[HTML]{CBD8FF}\textbf{41.3}} & \multicolumn{1}{c|}{\cellcolor[HTML]{CBD8FF}\textbf{39.8}} & \multicolumn{1}{c|}{\cellcolor[HTML]{CBD8FF}\textbf{41.8}} & \textbf{40.7} & \multicolumn{1}{c|}{\cellcolor[HTML]{CBD8FF}\textbf{42.4}} & \multicolumn{1}{c|}{\cellcolor[HTML]{CBD8FF}\textbf{33.8}} & \multicolumn{1}{c|}{\cellcolor[HTML]{CBD8FF}\textbf{48.3}} & \multicolumn{1}{c|}{\cellcolor[HTML]{CBD8FF}\textbf{46.3}} & \multicolumn{1}{c|}{\cellcolor[HTML]{CBD8FF}\textbf{48.9}} & \textbf{47.4} \\ \bottomrule
\end{tabular}

\label{tab:instance-seg}
\end{table*}

\begin{table}\scriptsize
\centering
\caption{Results of oriented object detection on DOTA dataset. Our SAQ-SAM shows consistent superiority in both comparison groups.}
\setlength{\tabcolsep}{2pt}
\begin{tabular}{@{}c|ccc|ccc|ccc@{}}
\toprule
 & \multicolumn{3}{c|}{SAM-B} & \multicolumn{3}{c|}{SAM-L} & \multicolumn{3}{c}{SAM-H} \\ \cmidrule(l){2-10} 
\multirow{-2}{*}{Method} & \multicolumn{1}{c|}{FP} & \multicolumn{1}{c|}{W6A6} & W4A4 & \multicolumn{1}{c|}{FP} & \multicolumn{1}{c|}{W6A6} & W4A4 & \multicolumn{1}{c|}{FP} & \multicolumn{1}{c|}{W6A6} & W4A4 \\ \midrule
\rowcolor[HTML]{F0FCEC} 
Percentile & \multicolumn{1}{c|}{\cellcolor[HTML]{F0FCEC}} & \multicolumn{1}{c|}{\cellcolor[HTML]{F0FCEC}10.35} & 2.62 & \multicolumn{1}{c|}{\cellcolor[HTML]{F0FCEC}} & \multicolumn{1}{c|}{\cellcolor[HTML]{F0FCEC}60.98} & 10.06 & \multicolumn{1}{c|}{\cellcolor[HTML]{F0FCEC}} & \multicolumn{1}{c|}{\cellcolor[HTML]{F0FCEC}60.08} & 14.55 \\
\rowcolor[HTML]{F0FCEC} 
OMSE & \multicolumn{1}{c|}{\cellcolor[HTML]{F0FCEC}} & \multicolumn{1}{c|}{\cellcolor[HTML]{F0FCEC}18.89} & 1.11 & \multicolumn{1}{c|}{\cellcolor[HTML]{F0FCEC}} & \multicolumn{1}{c|}{\cellcolor[HTML]{F0FCEC}61.55} & 23.50 & \multicolumn{1}{c|}{\cellcolor[HTML]{F0FCEC}} & \multicolumn{1}{c|}{\cellcolor[HTML]{F0FCEC}61.11} & 17.49 \\
\rowcolor[HTML]{F0FCEC} 
PTQ4SAM-S & \multicolumn{1}{c|}{\cellcolor[HTML]{F0FCEC}} & \multicolumn{1}{c|}{\cellcolor[HTML]{F0FCEC}19.90} & 0.58 & \multicolumn{1}{c|}{\cellcolor[HTML]{F0FCEC}} & \multicolumn{1}{c|}{\cellcolor[HTML]{F0FCEC}60.88} & 10.25 & \multicolumn{1}{c|}{\cellcolor[HTML]{F0FCEC}} & \multicolumn{1}{c|}{\cellcolor[HTML]{F0FCEC}62.99} & 40.14 \\
\rowcolor[HTML]{DCF5D3} 
SAQ-SAM$^{\star}$ & \multicolumn{1}{c|}{\multirow{-4}{*}{\cellcolor[HTML]{F0FCEC}63.93}} & \multicolumn{1}{c|}{\cellcolor[HTML]{DCF5D3}\textbf{51.84}} & \textbf{4.63} & \multicolumn{1}{c|}{\multirow{-4}{*}{\cellcolor[HTML]{F0FCEC}64.35}} & \multicolumn{1}{c|}{\cellcolor[HTML]{DCF5D3}\textbf{62.70}} & \textbf{27.83} & \multicolumn{1}{c|}{\multirow{-4}{*}{\cellcolor[HTML]{F0FCEC}64.49}} & \multicolumn{1}{c|}{\cellcolor[HTML]{DCF5D3}\textbf{63.35}} & \textbf{41.63} \\ \midrule
\rowcolor[HTML]{E7EDFF} 
AdaRound & \multicolumn{1}{c|}{\cellcolor[HTML]{E7EDFF}} & \multicolumn{1}{c|}{\cellcolor[HTML]{E7EDFF}34.05} & - & \multicolumn{1}{c|}{\cellcolor[HTML]{E7EDFF}} & \multicolumn{1}{c|}{\cellcolor[HTML]{E7EDFF}63.44} & 23.18 & \multicolumn{1}{c|}{\cellcolor[HTML]{E7EDFF}} & \multicolumn{1}{c|}{\cellcolor[HTML]{E7EDFF}62.73} & 24.45 \\
\rowcolor[HTML]{E7EDFF} 
BRECQ & \multicolumn{1}{c|}{\cellcolor[HTML]{E7EDFF}} & \multicolumn{1}{c|}{\cellcolor[HTML]{E7EDFF}34.40} & - & \multicolumn{1}{c|}{\cellcolor[HTML]{E7EDFF}} & \multicolumn{1}{c|}{\cellcolor[HTML]{E7EDFF}63.60} & 26.89 & \multicolumn{1}{c|}{\cellcolor[HTML]{E7EDFF}} & \multicolumn{1}{c|}{\cellcolor[HTML]{E7EDFF}62.58} & 25.98 \\
\rowcolor[HTML]{E7EDFF} 
QDrop & \multicolumn{1}{c|}{\cellcolor[HTML]{E7EDFF}} & \multicolumn{1}{c|}{\cellcolor[HTML]{E7EDFF}59.27} & 41.96 & \multicolumn{1}{c|}{\cellcolor[HTML]{E7EDFF}} & \multicolumn{1}{c|}{\cellcolor[HTML]{E7EDFF}63.86} & 50.11 & \multicolumn{1}{c|}{\cellcolor[HTML]{E7EDFF}} & \multicolumn{1}{c|}{\cellcolor[HTML]{E7EDFF}62.83} & 55.87 \\
\rowcolor[HTML]{E7EDFF} 
PTQ4SAM-L & \multicolumn{1}{c|}{\cellcolor[HTML]{E7EDFF}} & \multicolumn{1}{c|}{\cellcolor[HTML]{E7EDFF}60.33} & 44.18 & \multicolumn{1}{c|}{\cellcolor[HTML]{E7EDFF}} & \multicolumn{1}{c|}{\cellcolor[HTML]{E7EDFF}63.91} & 56.29 & \multicolumn{1}{c|}{\cellcolor[HTML]{E7EDFF}} & \multicolumn{1}{c|}{\cellcolor[HTML]{E7EDFF}64.36} & 56.01 \\
\rowcolor[HTML]{CBD8FF} 
SAQ-SAM & \multicolumn{1}{c|}{\multirow{-5}{*}{\cellcolor[HTML]{E7EDFF}63.93}} & \multicolumn{1}{c|}{\cellcolor[HTML]{CBD8FF}\textbf{60.35}} & \textbf{44.39} & \multicolumn{1}{c|}{\multirow{-5}{*}{\cellcolor[HTML]{E7EDFF}64.35}} & \multicolumn{1}{c|}{\cellcolor[HTML]{CBD8FF}\textbf{64.38}} & \textbf{58.03} & \multicolumn{1}{c|}{\multirow{-5}{*}{\cellcolor[HTML]{E7EDFF}64.49}} & \multicolumn{1}{c|}{\cellcolor[HTML]{CBD8FF}\textbf{64.58}} & \textbf{60.17} \\ \bottomrule
\end{tabular}
\label{tab:OOD}
\end{table}

\section{Experiments}

\subsection{Experimental Setup}
\textbf{Tasks and Datasets.} Our experiments include three prevalent tasks with SAM: instance segmentation, oriented object detection, and semantic segmentation. 
\begin{itemize}
\item In instance segmentation, various object detectors provide prompting boxes for SAM to obtain segmentation masks, including Faster R-CNN~\cite{faster-rcnn}, YOLOX~\cite{yolox}, H-Deformable-DETR~\cite{detrs}, and DINO~\cite{dino}, with MS-COCO~\cite{coco} as evaluation dataset.

\item In oriented object detection, with a detector generating horizontal boxes as prompts, SAM predicts masks that are subsequently transformed into resulting rotated boxes. We do experiment on DOTA~\cite{dota} dataset with FCOS~\cite{tian2019fcos} detector.

\item In the semantic segmentation, SAM is employed to enhance the quality of segmentation masks produced by traditional semantic segmentor.
We employ SegFormer~\cite{segformer} as the semantic branch and evaluate on ADE20K~\cite{ade20k} dataset.

\end{itemize}

\noindent\textbf{Implementation Details.} 
We maintain consistant experimental setting with baselines~\cite{ptq4sam} for fair comparision.
Specifically, the quantization scheme employs per-tensor asymmetric quantization for activations and per-channel asymmetric quantization for weights. The weights are calibrated using the MSE. The activation calibration set contains 32 images randomly sampled from the training dataset. For the PCC, we set threshold factor $\theta$ to 0.5. All QK activations (i.e., the inputs and outputs of Query Linear and Key Linear) are calibrated based on Attention Focus Overlap metric using the first sample.
For PAR, the image encoder employs per-stage learning, while the mask decoder employs per-layer learning with 2000 iterations. Exceptionally, the additional final cross-attention block learns 10000 iterations due to its large loss. In general, compared to PTQ4SAM settings, which carry per-block learning with 20000 iterations, our method requires much lower time cost. 

\noindent\textbf{Baseline Methods.} We take advanced PTQ4SAM as baseline, because both are dedicated to refining mask decoder quantization. For comprehensive comparison, we follow the PTQ4SAM settings and divide the methods into two groups according to whether there is a learning process. For statistic-based PTQ methods, we integrated PCC into PTQ4SAM-S, which is referred to as SAQ-SAM$^{\star}$. And for learning-based PTQ, we build on PTQ4SAM-L and modified the reconstruction method from QDrop~\cite{qdrop} to our PAR, with PCC as pre-calibration process.

\subsection{Performance Evaluation}
\noindent\textbf{Instance Segmentation.}
In the task of instance segmentation, 
as shown in Table \ref{tab:instance-seg}, our method consistently shows superior performance. Compared to baseline PTQ4SAM-S, SAQ-SAM$^{\star}$ achieves a nearly twofold improvement for 6-bit SAM-B, 
For SAM-L and SAM-H, the advantages are particularly evident in low-bit settings. For example, SAQ-SAM$^{\star}$ increases the mAP of 4-bit SAM-L from 23.1\% to 27.8\% with DINO. 
For reconstruction methods, compared to PTQ4SAM-L, our methods achieve significant improvement with lower computational costs. 
For example, SAQ-SAM improves the 4-bit SAM-B by 14.5\% mAP and achieves near-lossless accuracy for 4-bit SAM-L and SAM-H.

\noindent\textbf{Oriented Object Detection.}
In the oriented object detection task, as shown in Table \ref{tab:OOD}, our method also yields excellent performance, demonstrating generality for accurate segmentation and orientation of small targets. For example, in statistic-based setting, our SAQ-SAM$^{\star}$ boost 6-bit SAM-B performance to a usable level (from 19.9\% to 51.84\%). In learning-based setting, SAQ-SAM improves the mAP of 4-bit SAM-L from 56.29\% to 58.03\%.

\begin{table}\scriptsize
\centering
\caption{Results of semantic segmentation on ADE20K dataset. Our SAQ-SAM demonstrate effective in improving segmentation performance.}
\setlength{\tabcolsep}{5pt}
\begin{tabular}{@{}c|ccc|ccc@{}}
\toprule
 & \multicolumn{3}{c|}{SAM-B} & \multicolumn{3}{c}{SAM-L} \\ \cmidrule(l){2-7} 
\multirow{-2}{*}{Method} & \multicolumn{1}{c|}{FP} & \multicolumn{1}{c|}{W6A6} & W4A4 & \multicolumn{1}{c|}{FP} & \multicolumn{1}{c|}{W6A6} & W4A4 \\ \midrule
\rowcolor[HTML]{F0FCEC} 
PTQ4SAM-S & \multicolumn{1}{c|}{\cellcolor[HTML]{F0FCEC}} & \multicolumn{1}{c|}{\cellcolor[HTML]{F0FCEC}31.16} & 31.08 & \multicolumn{1}{c|}{\cellcolor[HTML]{F0FCEC}} & \multicolumn{1}{c|}{\cellcolor[HTML]{F0FCEC}33.48} & 22.64 \\
\rowcolor[HTML]{DCF5D3} 
SAQ-SAM$^{\star}$ & \multicolumn{1}{c|}{\multirow{-2}{*}{\cellcolor[HTML]{F0FCEC}33.15}} & \multicolumn{1}{c|}{\cellcolor[HTML]{DCF5D3}\textbf{32.90}} & \textbf{31.29} & \multicolumn{1}{c|}{\multirow{-2}{*}{\cellcolor[HTML]{F0FCEC}33.61}} & \multicolumn{1}{c|}{\cellcolor[HTML]{DCF5D3}\textbf{33.55}} & \textbf{25.64} \\ \midrule
\rowcolor[HTML]{E7EDFF} 
AdaRound & \multicolumn{1}{c|}{\cellcolor[HTML]{E7EDFF}} & \multicolumn{1}{c|}{\cellcolor[HTML]{E7EDFF}32.34} & 31.78 & \multicolumn{1}{c|}{\cellcolor[HTML]{E7EDFF}} & \multicolumn{1}{c|}{\cellcolor[HTML]{E7EDFF}32.99} & 31.97 \\
\rowcolor[HTML]{E7EDFF} 
BRECQ & \multicolumn{1}{c|}{\cellcolor[HTML]{E7EDFF}} & \multicolumn{1}{c|}{\cellcolor[HTML]{E7EDFF}32.27} & 31.78 & \multicolumn{1}{c|}{\cellcolor[HTML]{E7EDFF}} & \multicolumn{1}{c|}{\cellcolor[HTML]{E7EDFF}33.04} & 31.98 \\
\rowcolor[HTML]{E7EDFF} 
QDrop & \multicolumn{1}{c|}{\cellcolor[HTML]{E7EDFF}} & \multicolumn{1}{c|}{\cellcolor[HTML]{E7EDFF}32.57} & 31.79 & \multicolumn{1}{c|}{\cellcolor[HTML]{E7EDFF}} & \multicolumn{1}{c|}{\cellcolor[HTML]{E7EDFF}33.58} & 32.67 \\
\rowcolor[HTML]{E7EDFF} 
PTQ4SAM-L & \multicolumn{1}{c|}{\cellcolor[HTML]{E7EDFF}} & \multicolumn{1}{c|}{\cellcolor[HTML]{E7EDFF}32.65} & 31.85 & \multicolumn{1}{c|}{\cellcolor[HTML]{E7EDFF}} & \multicolumn{1}{c|}{\cellcolor[HTML]{E7EDFF}\textbf{33.66}} & 32.82 \\
\rowcolor[HTML]{CBD8FF} 
SAQ-SAM & \multicolumn{1}{c|}{\multirow{-5}{*}{\cellcolor[HTML]{E7EDFF}33.15}} & \multicolumn{1}{c|}{\cellcolor[HTML]{CBD8FF}\textbf{33.04}} & \textbf{32.53} & \multicolumn{1}{c|}{\multirow{-5}{*}{\cellcolor[HTML]{E7EDFF}33.61}} & \multicolumn{1}{c|}{\cellcolor[HTML]{CBD8FF}33.63} & \textbf{33.30} \\ \bottomrule
\end{tabular}
\label{tab:semantic-seg}
\end{table}

\noindent\textbf{Semantic Segmentation.}
In the task of semantic segmentation, the accuracy degradation caused by quantization is not significant due to the sophisticate post-processing steps. 
Nevertheless, as shown in Table \ref{tab:semantic-seg}, our methods also show superiority. For example, SAQ-SAM$^{\star}$ increases the accuracy of 6-bit SAM-B by 1.74\% mIoU,
even surpassing the learning-based PTQ4SAM-L.
And SAQ-SAM further boosts the performance close to FP.

\subsection{Ablation Study}
With PTQ4SAM-L as baseline, we explore the effectiveness of SAQ-SAM components. As shown in Table \ref{tab:main_ab}, while PAR outperforms the baseline, when PCC act as pre-calibration for PAR, the combination of both achieves the best performance. 
See Appendix \ref{sec:all_ab} for more analysis.

\begin{table}[h]\scriptsize
\centering
\caption{Ablation study of SAQ-SAM components.}
\setlength{\tabcolsep}{4pt}
\begin{tabular}{@{}c|cc|cc|cc|cc@{}}
\toprule
 & \multicolumn{2}{c|}{Method} & \multicolumn{2}{c|}{SAM-B} & \multicolumn{2}{c|}{SAM-L} & \multicolumn{2}{c}{SAM-H} \\ \cmidrule(l){2-9} 
\multirow{-2}{*}{Detector} & PAR & PCC & \multicolumn{1}{c|}{W6A6} & W4A4 & \multicolumn{1}{c|}{W6A6} & W4A4 & \multicolumn{1}{c|}{W6A6} & W4A4 \\ \midrule
 & \xmark & \xmark & \multicolumn{1}{c|}{\cellcolor[HTML]{E7EDFF}34.3} & \cellcolor[HTML]{E7EDFF}18.4 & \multicolumn{1}{c|}{\cellcolor[HTML]{E7EDFF}\textbf{40.3}} & \cellcolor[HTML]{E7EDFF}31.6 & \multicolumn{1}{c|}{\cellcolor[HTML]{E7EDFF}40.7} & \cellcolor[HTML]{E7EDFF}37.6 \\
 & \cmark & \xmark & \multicolumn{1}{c|}{\cellcolor[HTML]{E7EDFF}35.7} & \cellcolor[HTML]{E7EDFF}26.2 & \multicolumn{1}{c|}{\cellcolor[HTML]{E7EDFF}40.2} & \cellcolor[HTML]{E7EDFF}38.9 & \multicolumn{1}{c|}{\cellcolor[HTML]{E7EDFF}40.8} & \cellcolor[HTML]{E7EDFF}39.4 \\
\multirow{-3}{*}{YOLOX} & \cmark & \cmark & \multicolumn{1}{c|}{\cellcolor[HTML]{CBD8FF}\textbf{35.9}} & \cellcolor[HTML]{CBD8FF}\textbf{30.3} & \multicolumn{1}{c|}{\cellcolor[HTML]{CBD8FF}\textbf{40.3}} & \cellcolor[HTML]{CBD8FF}\textbf{39.0} & \multicolumn{1}{c|}{\cellcolor[HTML]{CBD8FF}\textbf{40.9}} & \cellcolor[HTML]{CBD8FF}\textbf{39.9} \\ \midrule
 & \xmark & \xmark & \multicolumn{1}{c|}{\cellcolor[HTML]{E7EDFF}40.4} & \cellcolor[HTML]{E7EDFF}14.4 & \multicolumn{1}{c|}{\cellcolor[HTML]{E7EDFF}\textbf{48.3}} & \cellcolor[HTML]{E7EDFF}36.6 & \multicolumn{1}{c|}{\cellcolor[HTML]{E7EDFF}48.7} & \cellcolor[HTML]{E7EDFF}43.9 \\
 & \cmark & \xmark & \multicolumn{1}{c|}{\cellcolor[HTML]{E7EDFF}42.3} & \cellcolor[HTML]{E7EDFF}30.2 & \multicolumn{1}{c|}{\cellcolor[HTML]{E7EDFF}\textbf{48.3}} & \cellcolor[HTML]{E7EDFF}46.1 & \multicolumn{1}{c|}{\cellcolor[HTML]{E7EDFF}48.8} & \cellcolor[HTML]{E7EDFF}\textbf{47.4} \\
\multirow{-3}{*}{DINO} & \cmark & \cmark & \multicolumn{1}{c|}{\cellcolor[HTML]{CBD8FF}\textbf{42.4}} & \cellcolor[HTML]{CBD8FF}\textbf{33.8} & \multicolumn{1}{c|}{\cellcolor[HTML]{CBD8FF}\textbf{48.3}} & \cellcolor[HTML]{CBD8FF}\textbf{46.3} & \multicolumn{1}{c|}{\cellcolor[HTML]{CBD8FF}\textbf{48.9}} & \cellcolor[HTML]{CBD8FF}\textbf{47.4} \\ \bottomrule
\end{tabular}

\label{tab:main_ab}
\end{table}

\begin{figure}
\centering
\includegraphics[width=1\linewidth]{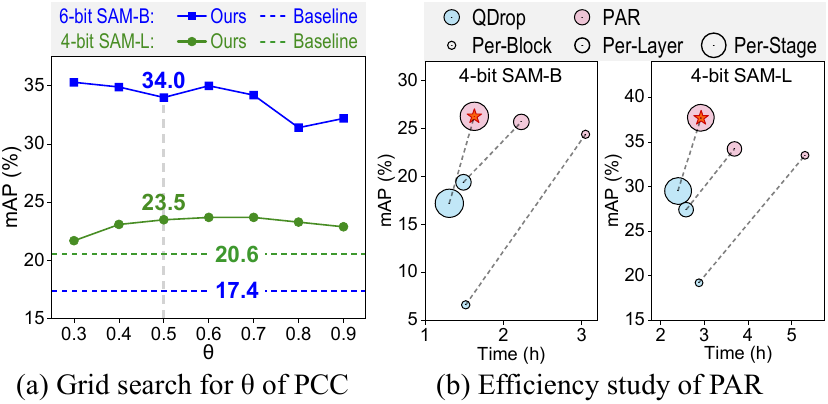}
\caption{
Ablation study of PCC and PAR. Figure (a) shows PCC is robust to the setting of $\theta$. Figure (b) shows per-stage PAR achieves optimal efficiency-accuracy trade-off.
}
\label{fig:ab}
\end{figure}

\noindent\textbf{Attention Focus Threshold Setting of PCC.}
In the proposed PCC module, we use the threshold factor $\theta$ to define the salient region of attention, which is typically set at 0.5. To explore the impact of this hyperparameter setting, we perform a grid
search on it. As shown in Figure \ref{fig:ab} (a), with changes in $\theta$, the performance of PCC consistently outperforms the baseline method, albeit with slight fluctuations, where 0.5 is a modest choice. Notably, our method demonstrates exceptional robustness to the setting of $\theta$.

\noindent\textbf{Efficiency-Accuracy Study of PAR.}
To further verify the superiority of PAR, 
we explore its time consumption and performance with different granularity of optimization units, and compare it with QDrop-based reconstruction.
As shown in Figure \ref{fig:ab},
per-stage scheme reduces the time cost of parameter learning, 
as well as boosting reconstruction performance,
where per-stage PAR obtains optimal efficiency-accuracy trade-off.
Furthermore, PAR performs better than QDrop under all granularity levels, highlighting the contribution of image-prompt interactions. 
The detailed experimental results can be found in the Appendix \ref{sec:rec_efficiency}.

PCC could serve as an orthogonal technique for combining with other advanced quantization methods. 
We present the improvement with RepQ-ViT and QDrop in the Appendix \ref{sec:RepQ-QDrop}.
Since our method does not introduce any additional model components, the compression and acceleration benefits of quantized SAM are consistent with that reported in PTQ4SAM paper. See Appendix \ref{sec:efficiency} for details.

\section{Conclusion}
In this paper, we propose SAQ-SAM, a post-training quantization framework for the SAM model. To address the issue of extreme outliers in mask decoder's QK activations, we introduce a semantically-preserved outlier clipping approach, which can effectively suppress significant outliers without smoothing or isolation.
Specifically, we determine the optimal clipping boundary by minimizing the attention focus deviation caused by quantization, overcoming the limitations of distribution-based methods.
Furthermore, we incorporate image-prompt interactions into reconstruction, which learns the correlation between visual features and prompt intent, thereby achieving alignment at both semantic and distributional levels. 
To improve the interaction efficiency, we also introduce a layer-skipping strategy and a stage-partitioning approach.
Extensive experiments show that our method significantly outperforms baseline methods, especially in low-bit scenarios. 


\section*{Acknowledgments}
This work was supported in part by the Strategic Priority Research Program of Chinese Academy of Sciences under Grant Number XDB1100000; 
in part by the National Natural Science Foundation of China under Grant Number 62276255;
in part by the Postdoctoral Fellowship Program of CPSF under Grant Number GZC20251175;
in part by the National Key Research and Development Program of China under Grant Number 2022ZD0119402;
in part by the Beijing Natural Science Foundation Haidian Original Innovation Joint Fund Project under Grant Number L232035, L222154.


\bibliography{aaai2026}

@inproceedings{win_attn,
  title={Exploring plain vision transformer backbones for object detection},
  author={Li, Yanghao and Mao, Hanzi and Girshick, Ross and He, Kaiming},
  booktitle={European conference on computer vision},
  pages={280--296},
  year={2022},
  organization={Springer}
}

@inproceedings{adaround,
  title={Up or down? adaptive rounding for post-training quantization},
  author={Nagel, Markus and Amjad, Rana Ali and Van Baalen, Mart and Louizos, Christos and Blankevoort, Tijmen},
  booktitle={International conference on machine learning},
  pages={7197--7206},
  year={2020},
  organization={PMLR}
}

@article{interactions-across-block,
  title={Interactions Across Blocks in Post-Training Quantization of Large Language Models},
  author={Shabanovi, Khasmamad and Wiest, Lukas and Golkov, Vladimir and Cremers, Daniel and Pfeil, Thomas},
  journal={arXiv preprint arXiv:2411.03934},
  year={2024}
}

@article{segformer,
  title={SegFormer: Simple and efficient design for semantic segmentation with transformers},
  author={Xie, Enze and Wang, Wenhai and Yu, Zhiding and Anandkumar, Anima and Alvarez, Jose M and Luo, Ping},
  journal={Advances in neural information processing systems},
  volume={34},
  pages={12077--12090},
  year={2021}
}

@article{dino,
  title={Dino: Detr with improved denoising anchor boxes for end-to-end object detection},
  author={Zhang, Hao and Li, Feng and Liu, Shilong and Zhang, Lei and Su, Hang and Zhu, Jun and Ni, Lionel M and Shum, Heung-Yeung},
  journal={arXiv preprint arXiv:2203.03605},
  year={2022}
}

@inproceedings{detrs,
  title={Detrs with hybrid matching},
  author={Jia, Ding and Yuan, Yuhui and He, Haodi and Wu, Xiaopei and Yu, Haojun and Lin, Weihong and Sun, Lei and Zhang, Chao and Hu, Han},
  booktitle={Proceedings of the IEEE/CVF conference on computer vision and pattern recognition},
  pages={19702--19712},
  year={2023}
}

@article{yolox,
  title={Yolox: Exceeding yolo series in 2021},
  author={Ge, Zheng and Liu, Songtao and Wang, Feng and Li, Zeming and Sun, Jian},
  journal={arXiv preprint arXiv:2107.08430},
  year={2021}
}

@article{faster-rcnn,
  title={Faster r-cnn: Towards real-time object detection with region proposal networks},
  author={Ren, Shaoqing and He, Kaiming and Girshick, Ross and Sun, Jian},
  journal={Advances in neural information processing systems},
  volume={28},
  year={2015}
}

@inproceedings{ade20k,
  title={Scene parsing through ade20k dataset},
  author={Zhou, Bolei and Zhao, Hang and Puig, Xavier and Fidler, Sanja and Barriuso, Adela and Torralba, Antonio},
  booktitle={Proceedings of the IEEE conference on computer vision and pattern recognition},
  pages={633--641},
  year={2017}
}

@inproceedings{coco,
  title={Microsoft coco: Common objects in context},
  author={Lin, Tsung-Yi and Maire, Michael and Belongie, Serge and Hays, James and Perona, Pietro and Ramanan, Deva and Doll{\'a}r, Piotr and Zitnick, C Lawrence},
  booktitle={Computer vision--ECCV 2014: 13th European conference, zurich, Switzerland, September 6-12, 2014, proceedings, part v 13},
  pages={740--755},
  year={2014},
  organization={Springer}
}

@misc{chen2023semantic,
    title = {Semantic Segment Anything},
    author = {Chen, Jiaqi and Yang, Zeyu and Zhang, Li},
    howpublished = {\url{https://github.com/fudan-zvg/Semantic-Segment-Anything}},
    year = {2023}
}

@inproceedings{pq-sam,
  title={PQ-SAM: Post-training Quantization for Segment Anything Model},
  author={Liu, Xiaoyu and Ding, Xin and Yu, Lei and Xi, Yuanyuan and Li, Wei and Tu, Zhijun and Hu, Jie and Chen, Hanting and Yin, Baoqun and Xiong, Zhiwei},
  booktitle={European Conference on Computer Vision},
  pages={420--437},
  year={2024},
  organization={Springer}
}

@inproceedings{ptq4sam,
  title={Ptq4sam: Post-training quantization for segment anything},
  author={Lv, Chengtao and Chen, Hong and Guo, Jinyang and Ding, Yifu and Liu, Xianglong},
  booktitle={Proceedings of the IEEE/CVF Conference on Computer Vision and Pattern Recognition},
  pages={15941--15951},
  year={2024}
}

@inproceedings{pd,
  title={Pd-quant: Post-training quantization based on prediction difference metric},
  author={Liu, Jiawei and Niu, Lin and Yuan, Zhihang and Yang, Dawei and Wang, Xinggang and Liu, Wenyu},
  booktitle={Proceedings of the IEEE/CVF Conference on Computer Vision and Pattern Recognition},
  pages={24427--24437},
  year={2023}
}

@article{qdrop,
  title={Qdrop: Randomly dropping quantization for extremely low-bit post-training quantization},
  author={Wei, Xiuying and Gong, Ruihao and Li, Yuhang and Liu, Xianglong and Yu, Fengwei},
  journal={arXiv preprint arXiv:2203.05740},
  year={2022}
}

@article{brecq,
  title={Brecq: Pushing the limit of post-training quantization by block reconstruction},
  author={Li, Yuhang and Gong, Ruihao and Tan, Xu and Yang, Yang and Hu, Peng and Zhang, Qi and Yu, Fengwei and Wang, Wei and Gu, Shi},
  journal={arXiv preprint arXiv:2102.05426},
  year={2021}
}

@article{ptqvit,
  title={Post-training quantization for vision transformer},
  author={Liu, Zhenhua and Wang, Yunhe and Han, Kai and Zhang, Wei and Ma, Siwei and Gao, Wen},
  journal={Advances in Neural Information Processing Systems},
  volume={34},
  pages={28092--28103},
  year={2021}
}

@article{omniquant,
  title={Omniquant: Omnidirectionally calibrated quantization for large language models},
  author={Shao, Wenqi and Chen, Mengzhao and Zhang, Zhaoyang and Xu, Peng and Zhao, Lirui and Li, Zhiqian and Zhang, Kaipeng and Gao, Peng and Qiao, Yu and Luo, Ping},
  journal={arXiv preprint arXiv:2308.13137},
  year={2023}
}

@inproceedings{smoothquant,
  title={Smoothquant: Accurate and efficient post-training quantization for large language models},
  author={Xiao, Guangxuan and Lin, Ji and Seznec, Mickael and Wu, Hao and Demouth, Julien and Han, Song},
  booktitle={International Conference on Machine Learning},
  pages={38087--38099},
  year={2023},
  organization={PMLR}
}

@inproceedings{efficientsam,
  title={Efficientsam: Leveraged masked image pretraining for efficient segment anything},
  author={Xiong, Yunyang and Varadarajan, Bala and Wu, Lemeng and Xiang, Xiaoyu and Xiao, Fanyi and Zhu, Chenchen and Dai, Xiaoliang and Wang, Dilin and Sun, Fei and Iandola, Forrest and others},
  booktitle={Proceedings of the IEEE/CVF Conference on Computer Vision and Pattern Recognition},
  pages={16111--16121},
  year={2024}
}

@article{fastsam,
  title={Fast segment anything},
  author={Zhao, Xu and Ding, Wenchao and An, Yongqi and Du, Yinglong and Yu, Tao and Li, Min and Tang, Ming and Wang, Jinqiao},
  journal={arXiv preprint arXiv:2306.12156},
  year={2023}
}

@article{mobilesam,
  title={Faster segment anything: Towards lightweight sam for mobile applications},
  author={Zhang, Chaoning and Han, Dongshen and Qiao, Yu and Kim, Jung Uk and Bae, Sung-Ho and Lee, Seungkyu and Hong, Choong Seon},
  journal={arXiv preprint arXiv:2306.14289},
  year={2023}
}

@inproceedings{kirillov2023segment,
  title={Segment anything},
  author={Kirillov, Alexander and Mintun, Eric and Ravi, Nikhila and Mao, Hanzi and Rolland, Chloe and Gustafson, Laura and Xiao, Tete and Whitehead, Spencer and Berg, Alexander C and Lo, Wan-Yen and others},
  booktitle={Proceedings of the IEEE/CVF international conference on computer vision},
  pages={4015--4026},
  year={2023}
}

@article{sammed2d,
  title={Sam-med2d},
  author={Cheng, Junlong and Ye, Jin and Deng, Zhongying and Chen, Jianpin and Li, Tianbin and Wang, Haoyu and Su, Yanzhou and Huang, Ziyan and Chen, Jilong and Jiang, Lei and others},
  journal={arXiv preprint arXiv:2308.16184},
  year={2023}
}

@article{MedSAM,
  title={Segment anything in medical images},
  author={Ma, Jun and He, Yuting and Li, Feifei and Han, Lin and You, Chenyu and Wang, Bo},
  journal={Nature Communications},
  volume={15},
  number={1},
  pages={654},
  year={2024},
  publisher={Nature Publishing Group UK London}
}

@article{vaswani2017attention,
  title={Attention is all you need},
  author={Vaswani, Ashish and Shazeer, Noam and Parmar, Niki and Uszkoreit, Jakob and Jones, Llion and Gomez, Aidan N and Kaiser, {\L}ukasz and Polosukhin, Illia},
  journal={Advances in neural information processing systems},
  volume={30},
  year={2017}
}

@article{gholami2021survey,
  title={A survey of quantization methods for efficient neural network inference},
  author={Gholami, Amir and Kim, Sehoon and Dong, Zhen and Yao, Zhewei and Mahoney, Michael W and Keutzer, Kurt},
  journal={arXiv preprint arXiv:2103.13630},
  year={2021}
}

@article{lin2024awq,
  title={Awq: Activation-aware weight quantization for on-device llm compression and acceleration},
  author={Lin, Ji and Tang, Jiaming and Tang, Haotian and Yang, Shang and Chen, Wei-Ming and Wang, Wei-Chen and Xiao, Guangxuan and Dang, Xingyu and Gan, Chuang and Han, Song},
  journal={Proceedings of Machine Learning and Systems},
  volume={6},
  pages={87--100},
  year={2024}
}

@inproceedings{zhang2024efficientvit,
  title={Efficientvit-sam: Accelerated segment anything model without performance loss},
  author={Zhang, Zhuoyang and Cai, Han and Han, Song},
  booktitle={Proceedings of the IEEE/CVF Conference on Computer Vision and Pattern Recognition},
  pages={7859--7863},
  year={2024}
}

@article{chen2023slimsam,
  title={SlimSAM: 0.1\% Data Makes Segment Anything Slim},
  author={Chen, Zigeng and Fang, Gongfan and Ma, Xinyin and Wang, Xinchao},
  journal={arXiv preprint arXiv:2312.05284},
  year={2023}
}

@article{ravi2024sam,
  title={Sam 2: Segment anything in images and videos},
  author={Ravi, Nikhila and Gabeur, Valentin and Hu, Yuan-Ting and Hu, Ronghang and Ryali, Chaitanya and Ma, Tengyu and Khedr, Haitham and R{\"a}dle, Roman and Rolland, Chloe and Gustafson, Laura and others},
  journal={arXiv preprint arXiv:2408.00714},
  year={2024}
}

@article{choi2018pact,
  title={Pact: Parameterized clipping activation for quantized neural networks},
  author={Choi, Jungwook and Wang, Zhuo and Venkataramani, Swagath and Chuang, Pierce I-Jen and Srinivasan, Vijayalakshmi and Gopalakrishnan, Kailash},
  journal={arXiv preprint arXiv:1805.06085},
  year={2018}
}

@inproceedings{tian2019fcos,
  title={Fcos: Fully convolutional one-stage object detection},
  author={Tian, Zhi and Shen, Chunhua and Chen, Hao and He, Tong},
  booktitle={Proceedings of the IEEE/CVF international conference on computer vision},
  pages={9627--9636},
  year={2019}
}

@article{yu2023h2rbox,
  title={H2RBox-v2: Incorporating symmetry for boosting horizontal box supervised oriented object detection},
  author={Yu, Yi and Yang, Xue and Li, Qingyun and Zhou, Yue and Da, Feipeng and Yan, Junchi},
  journal={Advances in Neural Information Processing Systems},
  volume={36},
  pages={59137--59150},
  year={2023}
}

@ARTICLE{dota,
  author={Ding, Jian and Xue, Nan and Xia, Gui-Song and Bai, Xiang and Yang, Wen and Yang, Michael and Belongie, Serge and Luo, Jiebo and Datcu, Mihai and Pelillo, Marcello and Zhang, Liangpei},
  journal={IEEE Transactions on Pattern Analysis and Machine Intelligence},
  title={Object Detection in Aerial Images: A Large-Scale Benchmark and Challenges},
  year={2021},
  volume={},
  number={},
  pages={1-1},
  doi={10.1109/TPAMI.2021.3117983}}

@article{li2025sam_ano,
  title={A SAM-guided two-stream lightweight model for anomaly detection},
  author={Li, Chenghao and Qi, Lei and Geng, Xin},
  journal={ACM Transactions on Multimedia Computing, Communications and Applications},
  volume={21},
  number={2},
  pages={1--23},
  year={2025},
  publisher={ACM New York, NY}
}

@article{chen2024rsprompter,
  title={RSPrompter: Learning to prompt for remote sensing instance segmentation based on visual foundation model},
  author={Chen, Keyan and Liu, Chenyang and Chen, Hao and Zhang, Haotian and Li, Wenyuan and Zou, Zhengxia and Shi, Zhenwei},
  journal={IEEE Transactions on Geoscience and Remote Sensing},
  volume={62},
  pages={1--17},
  year={2024},
  publisher={IEEE}
}

@inproceedings{liu2025cachequant,
  title={Cachequant: Comprehensively accelerated diffusion models},
  author={Liu, Xuewen and Li, Zhikai and Gu, Qingyi},
  booktitle={Proceedings of the Computer Vision and Pattern Recognition Conference},
  pages={23269--23280},
  year={2025}
}

@article{liu2024dilatequant,
  title={Dilatequant: Accurate and efficient diffusion quantization via weight dilation},
  author={Liu, Xuewen and Li, Zhikai and Gu, Qingyi},
  journal={arXiv preprint arXiv:2409.14307},
  year={2024}
}

@article{li2024htq,
  title={HTQ: Exploring the High-Dimensional Trade-Off of mixed-precision quantization},
  author={Li, Zhikai and Long, Xianlei and Xiao, Junrui and Gu, Qingyi},
  journal={Pattern Recognition},
  volume={156},
  pages={110788},
  year={2024},
  publisher={Elsevier}
}

@article{li2024privacy,
  title={Privacy-Preserving SAM Quantization for Efficient Edge Intelligence in Healthcare},
  author={Li, Zhikai and Zhang, Jing and Gu, Qingyi},
  journal={arXiv preprint arXiv:2410.01813},
  year={2024}
}

@article{xiao2024binaryvit,
  title={Binaryvit: Towards efficient and accurate binary vision transformers},
  author={Xiao, Junrui and Li, Zhikai and Li, Jianquan and Yang, Lianwei and Gu, Qingyi},
  journal={IEEE Transactions on Circuits and Systems for Video Technology},
  year={2024},
  publisher={IEEE}
}

@article{repquant,
  title={Repquant: Towards accurate post-training quantization of large transformer models via scale reparameterization},
  author={Li, Zhikai and Liu, Xuewen and Zhang, Jing and Gu, Qingyi},
  journal={arXiv preprint arXiv:2402.05628},
  year={2024}
}

@article{liu2024eda,
  title={EDA-DM: Enhanced Distribution Alignment for Post-Training Quantization of Diffusion Models},
  author={Liu, Xuewen and Li, Zhikai and Xiao, Junrui and Gu, Qingyi},
  journal={arXiv preprint arXiv:2401.04585},
  year={2024}
}

@article{li2023qft,
  title={Qft: Quantized full-parameter tuning of llms with affordable resources},
  author={Li, Zhikai and Liu, Xiaoxuan and Zhu, Banghua and Dong, Zhen and Gu, Qingyi and Keutzer, Kurt},
  journal={arXiv preprint arXiv:2310.07147},
  year={2023}
}

@article{li2023psaq,
  title={Psaq-vit v2: Toward accurate and general data-free quantization for vision transformers},
  author={Li, Zhikai and Chen, Mengjuan and Xiao, Junrui and Gu, Qingyi},
  journal={IEEE Transactions on Neural Networks and Learning Systems},
  year={2023},
  publisher={IEEE}
}

@inproceedings{repq-vit,
  title={Repq-vit: Scale reparameterization for post-training quantization of vision transformers},
  author={Li, Zhikai and Xiao, Junrui and Yang, Lianwei and Gu, Qingyi},
  booktitle={Proceedings of the IEEE/CVF International Conference on Computer Vision},
  pages={17227--17236},
  year={2023}
}

@inproceedings{li2023vit,
  title={I-vit: Integer-only quantization for efficient vision transformer inference},
  author={Li, Zhikai and Gu, Qingyi},
  booktitle={Proceedings of the IEEE/CVF International Conference on Computer Vision},
  pages={17065--17075},
  year={2023}
}

@article{li2022dual,
  title={Dual-discriminator adversarial framework for data-free quantization},
  author={Li, Zhikai and Ma, Liping and Long, Xianlei and Xiao, Junrui and Gu, Qingyi},
  journal={Neurocomputing},
  volume={511},
  pages={67--77},
  year={2022},
  publisher={Elsevier}
}

@inproceedings{li2022patch,
  title={Patch similarity aware data-free quantization for vision transformers},
  author={Li, Zhikai and Ma, Liping and Chen, Mengjuan and Xiao, Junrui and Gu, Qingyi},
  booktitle={European conference on computer vision},
  pages={154--170},
  year={2022},
  organization={Springer}
}

\newpage

\appendix

\section{Compression and Speedup Effects.}

\subsection{The Efficiency Gains of SAM Quantization}
\label{sec:efficiency}
Our SAQ-SAM does not introduce any additional model parameters of components for the quantized SAM models, so the compression and acceleration benefits are exactly consistent with those reported in the baseline paper (e.g., PTQ4SAM~\cite{ptq4sam}), as shown in Figure\ref{tab:acceleration}. FLOPs is used to measure the acceleration effect.
As models scale up, the acceleration ratio becomes more significant, while the performance drop becomes less.

\begin{table}[ht]\small
\centering
\setlength{\tabcolsep}{5pt}
\caption{
Acceleration ratio and storage reduction ratio of SAM quantization. 
}
\begin{tabular}{@{}ccccc@{}}
\toprule
Model & Params. (M) & \begin{tabular}[c]{@{}c@{}}Prec.\\ (W/A)\end{tabular} & \begin{tabular}[c]{@{}c@{}}Acceleration\\ Ratio\end{tabular} & \begin{tabular}[c]{@{}c@{}}Storage\\ Reduction\end{tabular} \\ \midrule
\multirow{2}{*}{SAM-B} & \multirow{2}{*}{90} & 6/6 & 2.96 $\times$ & 5.3 $\times$ \\
 &  & 4/4 & 3.48 $\times$ & 8.0 $\times$ \\ \midrule
\multirow{2}{*}{SAM-L} & \multirow{2}{*}{311} & 6/6 & 3.88 $\times$ & 5.3 $\times$ \\
 &  & 4/4 & 4.98 $\times$ & 8.0 $\times$ \\ \midrule
\multirow{2}{*}{SAM-H} & \multirow{2}{*}{636} & 6/6 & 4.40 $\times$ & 5.3 $\times$ \\
 &  & 4/4 & 5.00 $\times$ & 8.0 $\times$ \\ \bottomrule
\end{tabular}
\label{tab:acceleration}
\end{table}

\section{Discussion on PCC}

\subsection{More Implementation details.}
PCC addresses the issue of extreme outliers by maintaining semantic perceptual consistency in the attention module, where QK activations (e.g., the inputs and outputs of QK linear) are manipulated to achieve this purpose. Based on this, the proposed metric is specifically applied to QK activations, while other activations continue to be calibrated using the MSE metric. 
Given the characteristic of extremely-distributed activations, i.e., the presence of a small number of huge outliers hundreds of times greater than the normal values, we adopt the following design in our implementation: 
(1) To improve efficiency, we use a percentile-based method to search for optimal clipping range;
(2) To ensure the quantization model to adapt to drastic changes in activation values, the QK activations are first calibrated and kept in quantization state, and then other activations are calibrated with OMSE.
This strategy enables a balanced trade-off between distributional alignment and semantic fidelity across the model. By maintaining this balance, we prevent the risk of overemphasizing semantic alignment, which could otherwise lead to distributional collapse and degrade overall model performance.

\subsection{Extreme Distribution Issue and Validity Discussion for PCC.}
\label{sec:more_distribution}

\begin{figure}[ht]
\centering
\caption{
Visualization examples of query inputs distribution in the mask decoder. 
}
\includegraphics[width=0.9\linewidth]{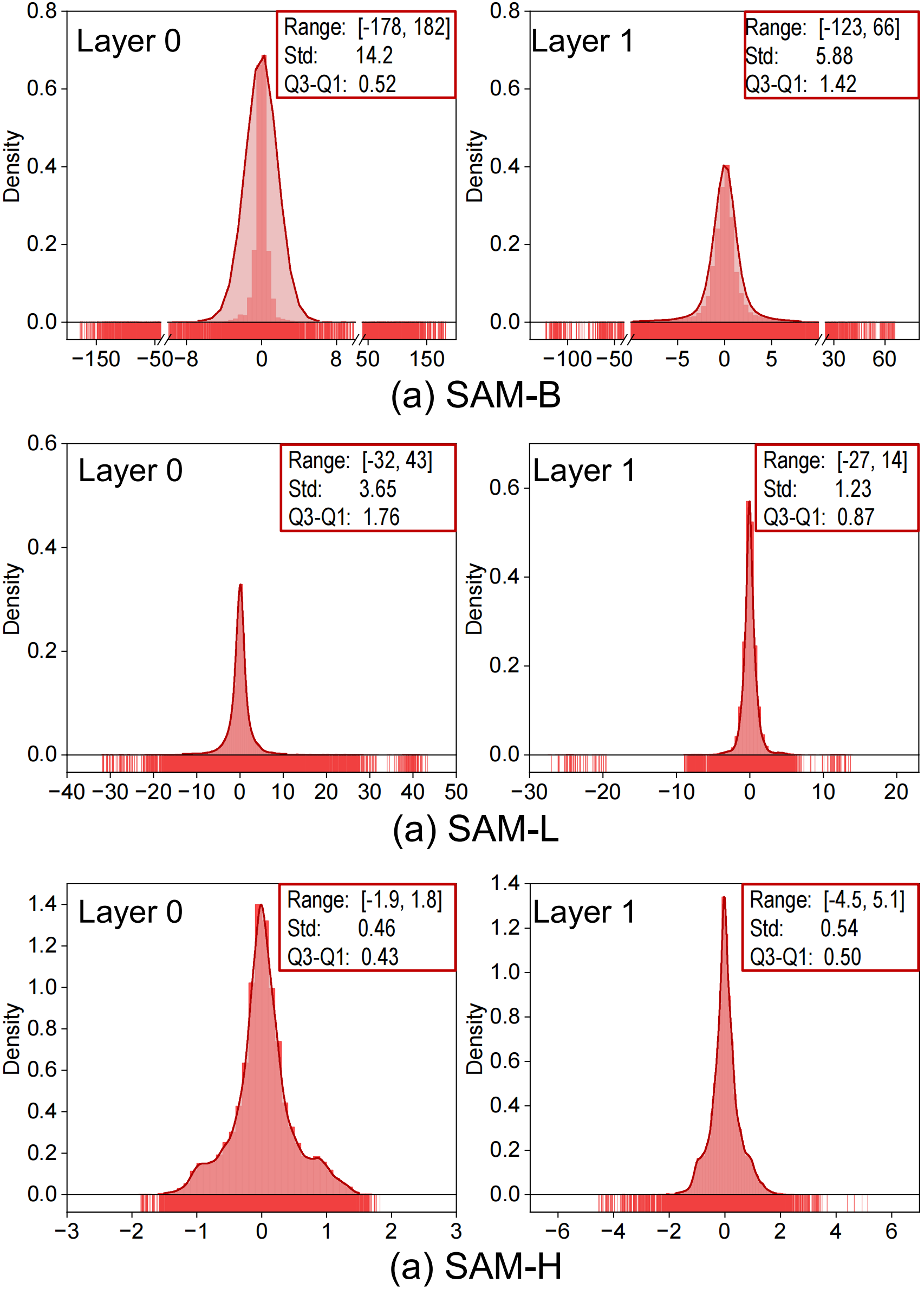}
\label{fig:distribution_more}
\end{figure}

We provide more visualizations of activation distribution in the mask decoder, as shown in Figure \ref{fig:distribution_more}. These activations are hooked from the query inputs of token-to-image cross-attention in the mask decoder, which exhibit most pronounced extreme outliers. We can observe that with larger model sizes, the issue of extreme outliers is mitigated. Nevertheless, as shown in the experiment results in the main section, our PCC still achieves performance gains compared with other distribution-based methods, demonstrating the validity of the proposed semantically aligned clipping. 

\subsection{Combining PCC with Other Advanced PTQ methods for ViT}
\label{sec:RepQ-QDrop}
Our PCC is dedicated to solving the quantization challenges of mask decoders. Currently, there are also some advanced ViT PTQ methods (e.g., RepQ-ViT) that, although structurally incompatible with mask decoder, can improve the quantization of image encoder. Therefore, PCC can be used in combination with these methods to achieve better performance without learning process. In addition, for traditional quantization reconstruction methods (e.g., QDrop), PCC can also provide better optimization initial values than MinMax, thus accelerating convergence. Experimental results for instance segmentation task are shown in Table \ref{tab:RepQ-Qdrop}, with the YOLOX detector on the COCO dataset. PCC achieves significant performance improvements compared to PTQ4SAM, which also focuses on improving mask decoder quantization.

\begin{table}[ht]\scriptsize
\centering
\caption{
The results of combining PCC with RepQ-ViT or QDrop. For RepQ-ViT, it is used for image encoder, which adopt standard ViT architecture, while PCC is used for mask decoders, which have complex structures that cannot be adapted to conventional outlier processing methods such as rotation and transformation.
}
\setlength{\tabcolsep}{3pt}
\begin{tabular}{@{}c|ccc|ccc@{}}
\toprule
 & \multicolumn{3}{c|}{SAM-B} & \multicolumn{3}{c}{SAM-L} \\ \cmidrule(l){2-7} 
\multirow{-2}{*}{Method} & \multicolumn{1}{c|}{FP} & \multicolumn{1}{c|}{W6A6} & W4A4 & \multicolumn{1}{c|}{FP} & \multicolumn{1}{c|}{W6A6} & W4A4 \\ \midrule
\rowcolor[HTML]{F0FCEC} 
RepQ & \multicolumn{1}{c|}{\cellcolor[HTML]{F0FCEC}} & \multicolumn{1}{c|}{\cellcolor[HTML]{F0FCEC}14.5} & 0.4 & \multicolumn{1}{c|}{\cellcolor[HTML]{F0FCEC}} & \multicolumn{1}{c|}{\cellcolor[HTML]{F0FCEC}38.8} & 13.7 \\
\rowcolor[HTML]{F0FCEC} 
RepQ+PTQ4SAM & \multicolumn{1}{c|}{\cellcolor[HTML]{F0FCEC}} & \multicolumn{1}{c|}{\cellcolor[HTML]{F0FCEC}19.3} & 0.3 & \multicolumn{1}{c|}{\cellcolor[HTML]{F0FCEC}} & \multicolumn{1}{c|}{\cellcolor[HTML]{F0FCEC}39.9} & 25.5 \\
\rowcolor[HTML]{DCF5D3} 
RepQ+PCC & \multicolumn{1}{c|}{\multirow{-3}{*}{\cellcolor[HTML]{F0FCEC}37.0}} & \multicolumn{1}{c|}{\cellcolor[HTML]{DCF5D3}\textbf{33.8}} & \textbf{1.2} & \multicolumn{1}{c|}{\multirow{-3}{*}{\cellcolor[HTML]{F0FCEC}40.4}} & \multicolumn{1}{c|}{\cellcolor[HTML]{DCF5D3}\textbf{40.0}} & \textbf{29.0} \\ \midrule
\rowcolor[HTML]{E7EDFF} 
QDrop & \multicolumn{1}{c|}{\cellcolor[HTML]{E7EDFF}} & \multicolumn{1}{c|}{\cellcolor[HTML]{E7EDFF}33.6} & 13.3 & \multicolumn{1}{c|}{\cellcolor[HTML]{E7EDFF}} & \multicolumn{1}{c|}{\cellcolor[HTML]{E7EDFF}39.7} & 25.3 \\
\rowcolor[HTML]{E7EDFF} 
QDrop+PTQ4SAM & \multicolumn{1}{c|}{\cellcolor[HTML]{E7EDFF}} & \multicolumn{1}{c|}{\cellcolor[HTML]{E7EDFF}34.3} & 18.4 & \multicolumn{1}{c|}{\cellcolor[HTML]{E7EDFF}} & \multicolumn{1}{c|}{\cellcolor[HTML]{E7EDFF}\textbf{40.3}} & 31.6 \\
\rowcolor[HTML]{CBD8FF} 
QDrop+PCC & \multicolumn{1}{c|}{\multirow{-3}{*}{\cellcolor[HTML]{E7EDFF}37.0}} & \multicolumn{1}{c|}{\cellcolor[HTML]{CBD8FF}\textbf{35.1}} & \textbf{21.2} & \multicolumn{1}{c|}{\multirow{-3}{*}{\cellcolor[HTML]{E7EDFF}40.4}} & \multicolumn{1}{c|}{\cellcolor[HTML]{CBD8FF}\textbf{40.3}} & \textbf{37.1} \\ \bottomrule
\end{tabular}
\label{tab:RepQ-Qdrop}
\end{table}

\section{Discussion on PAR.}
\subsection{Detailed Time-Accuracy Comparison of Reconstruction Methods}
\label{sec:rec_efficiency}
To investigate the additional time overhead introduced by the PAR interaction module, we conducted a detailed comparison of the time consumption of different reconstruction granularities (e.g., per-stage, per-layer and per-block) and different reconstruction methods (e.g., QDrop and PAR), and provided the corresponding performance results, as shown in Table \ref{tab:recon_time_cost}. To ensure a fair comparison, the number of iterations was kept consistent at 2000.

We provide more intuitive plots in the main text, and specific figures are given here. For example, compared to traditional reconstruction settings (per-block QDrop), our method (per-stage PAR) achieves significant performance improvement (e.g., from 6.6\% to 26.3\% with SAM-B), with slightly more time cost (e.g., from 1.52 h to 1.63 h with SAM-B). We can also observe that PAR exhibits performance advantages across all granularity levels, despite higher time consumption.

\begin{table}[ht]\scriptsize
\centering
\setlength{\tabcolsep}{5pt}
\caption{
The comparison of time consumption and performance with different reconstruction methods in various optimization units (Granularity).
}
\begin{tabular}{@{}cccccc@{}}
\toprule
Prec. & Model & Granularity & Method & mAP (\%) & Time (h) \\ \midrule
\multirow{6}{*}{W4A4} & \multirow{6}{*}{SAM-B} & \multirow{2}{*}{Per-Stage} & Qdrop & 17.2 & 1.31 \\
 &  &  & PAR & 26.3 & 1.63 \\ \cmidrule(l){3-6} 
 &  & \multirow{2}{*}{Per-Layer} & Qdrop & 19.4 & 1.49 \\
 &  &  & PAR & 25.7 & 2.23 \\ \cmidrule(l){3-6} 
 &  & \multirow{2}{*}{Per-Block} & Qdrop & 6.6 & 1.52 \\
 &  &  & PAR & 24.4 & 3.05 \\ \midrule
\multirow{6}{*}{W4A4} & \multirow{6}{*}{SAM-L} & \multirow{2}{*}{Per-Stage} & Qdrop & 29.5 & 2.40 \\
 &  &  & PAR & 37.7 & 2.93 \\ \cmidrule(l){3-6} 
 &  & \multirow{2}{*}{Per-Layer} & Qdrop & 27.4 & 2.59 \\
 &  &  & PAR & 34.2 & 3.69 \\ \cmidrule(l){3-6} 
 &  & \multirow{2}{*}{Per-Block} & Qdrop & 19.2 & 2.88 \\
 &  &  & PAR & 33.5 & 5.31 \\ \bottomrule
\end{tabular}
\label{tab:recon_time_cost}
\end{table}

\subsection{More Ablation Results.}
\label{sec:all_ab}
Due to space limitations, in the main text, we only present the detailed results of the ablation experiment with two detectors. Here, Table \ref{tab:all_ab} provides more comprehensive results, showing that PCC can provide better initial quantification parameters for PAR, especially for 4-bit SAM-B. For larger models, PAR can achieve satisfactory performance without PCC. Nevertheless, other experimental results in the main text (Table \ref{tab:instance-seg}, \ref{tab:OOD}, \ref{tab:semantic-seg}) indicate that PCC can achieve surprising performance improvement without compute-intensive learning, making it more friendly for resource-constrained devices.

\begin{table}[ht]\scriptsize
\centering
\setlength{\tabcolsep}{2pt}
\caption{
Ablation study results for SAQ-SAM with more detectors.
}
\begin{tabular}{@{}c|cc|cc|cc|cc@{}}
\toprule
 & \multicolumn{2}{c|}{Method} & \multicolumn{2}{c|}{SAM-B} & \multicolumn{2}{c|}{SAM-L} & \multicolumn{2}{c}{SAM-H} \\ \cmidrule(l){2-9} 
\multirow{-2}{*}{Detector} & PAR & PCC & \multicolumn{1}{c|}{W6A6} & W4A4 & \multicolumn{1}{c|}{W6A6} & W4A4 & \multicolumn{1}{c|}{W6A6} & W4A4 \\ \midrule
 & \xmark & \xmark & \multicolumn{1}{c|}{\cellcolor[HTML]{E7EDFF}34.3} & \cellcolor[HTML]{E7EDFF}18.4 & \multicolumn{1}{c|}{\cellcolor[HTML]{E7EDFF}\textbf{40.3}} & \cellcolor[HTML]{E7EDFF}31.6 & \multicolumn{1}{c|}{\cellcolor[HTML]{E7EDFF}40.7} & \cellcolor[HTML]{E7EDFF}37.6 \\
 & \cmark & \xmark & \multicolumn{1}{c|}{\cellcolor[HTML]{E7EDFF}35.7} & \cellcolor[HTML]{E7EDFF}26.2 & \multicolumn{1}{c|}{\cellcolor[HTML]{E7EDFF}40.2} & \cellcolor[HTML]{E7EDFF}38.9 & \multicolumn{1}{c|}{\cellcolor[HTML]{E7EDFF}40.8} & \cellcolor[HTML]{E7EDFF}39.4 \\
\multirow{-3}{*}{YOLOX} & \cmark & \cmark & \multicolumn{1}{c|}{\cellcolor[HTML]{CBD8FF}\textbf{35.9}} & \cellcolor[HTML]{CBD8FF}\textbf{30.3} & \multicolumn{1}{c|}{\cellcolor[HTML]{CBD8FF}\textbf{40.3}} & \cellcolor[HTML]{CBD8FF}\textbf{39.0} & \multicolumn{1}{c|}{\cellcolor[HTML]{CBD8FF}\textbf{40.9}} & \cellcolor[HTML]{CBD8FF}\textbf{39.9} \\ \midrule
 & \xmark & \xmark & \multicolumn{1}{c|}{\cellcolor[HTML]{E7EDFF}40.4} & \cellcolor[HTML]{E7EDFF}14.4 & \multicolumn{1}{c|}{\cellcolor[HTML]{E7EDFF}\textbf{48.3}} & \cellcolor[HTML]{E7EDFF}36.6 & \multicolumn{1}{c|}{\cellcolor[HTML]{E7EDFF}48.7} & \cellcolor[HTML]{E7EDFF}43.9 \\
 & \cmark & \xmark & \multicolumn{1}{c|}{\cellcolor[HTML]{E7EDFF}42.3} & \cellcolor[HTML]{E7EDFF}30.2 & \multicolumn{1}{c|}{\cellcolor[HTML]{E7EDFF}\textbf{48.3}} & \cellcolor[HTML]{E7EDFF}46.1 & \multicolumn{1}{c|}{\cellcolor[HTML]{E7EDFF}48.8} & \cellcolor[HTML]{E7EDFF}\textbf{47.4} \\
\multirow{-3}{*}{DINO} & \cmark & \cmark & \multicolumn{1}{c|}{\cellcolor[HTML]{CBD8FF}\textbf{42.4}} & \cellcolor[HTML]{CBD8FF}\textbf{33.8} & \multicolumn{1}{c|}{\cellcolor[HTML]{CBD8FF}\textbf{48.3}} & \cellcolor[HTML]{CBD8FF}\textbf{46.3} & \multicolumn{1}{c|}{\cellcolor[HTML]{CBD8FF}\textbf{48.9}} & \cellcolor[HTML]{CBD8FF}\textbf{47.4} \\ \midrule
 & \xmark & \xmark & \multicolumn{1}{c|}{\cellcolor[HTML]{E7EDFF}30.3} & \cellcolor[HTML]{E7EDFF}16.0 & \multicolumn{1}{c|}{\cellcolor[HTML]{E7EDFF}35.8} & \cellcolor[HTML]{E7EDFF}28.7 & \multicolumn{1}{c|}{\cellcolor[HTML]{E7EDFF}36.5} & \cellcolor[HTML]{E7EDFF}33.5 \\
 & \cmark & \xmark & \multicolumn{1}{c|}{\cellcolor[HTML]{E7EDFF}31.9} & \cellcolor[HTML]{E7EDFF}23.2 & \multicolumn{1}{c|}{\cellcolor[HTML]{E7EDFF}35.8} & \cellcolor[HTML]{E7EDFF}\textbf{34.6} & \multicolumn{1}{c|}{\cellcolor[HTML]{E7EDFF}\textbf{36.7}} & \cellcolor[HTML]{E7EDFF}\textbf{35.9} \\
\multirow{-3}{*}{Faster R-CNN} & \cmark & \cmark & \multicolumn{1}{c|}{\cellcolor[HTML]{CBD8FF}\textbf{32.0}} & \cellcolor[HTML]{CBD8FF}\textbf{27.7} & \multicolumn{1}{c|}{\cellcolor[HTML]{CBD8FF}\textbf{35.9}} & \cellcolor[HTML]{CBD8FF}34.5 & \multicolumn{1}{c|}{\cellcolor[HTML]{CBD8FF}36.6} & \cellcolor[HTML]{CBD8FF}\textbf{35.9} \\ \midrule
 & \xmark & \xmark & \multicolumn{1}{c|}{\cellcolor[HTML]{E7EDFF}35.1} & \cellcolor[HTML]{E7EDFF}17.3 & \multicolumn{1}{c|}{\cellcolor[HTML]{E7EDFF}41.2} & \cellcolor[HTML]{E7EDFF}32.1 & \multicolumn{1}{c|}{\cellcolor[HTML]{E7EDFF}41.6} & \cellcolor[HTML]{E7EDFF}38.4 \\
 & \cmark & \xmark & \multicolumn{1}{c|}{\cellcolor[HTML]{E7EDFF}37.0} & \cellcolor[HTML]{E7EDFF}29.4 & \multicolumn{1}{c|}{\cellcolor[HTML]{E7EDFF}41.2} & \cellcolor[HTML]{E7EDFF}39.6 & \multicolumn{1}{c|}{\cellcolor[HTML]{E7EDFF}\textbf{41.9}} & \cellcolor[HTML]{E7EDFF}\textbf{40.8} \\
\multirow{-3}{*}{H-Deformable-DETR} & \cmark & \cmark & \multicolumn{1}{c|}{\cellcolor[HTML]{CBD8FF}\textbf{37.1}} & \cellcolor[HTML]{CBD8FF}\textbf{31.8} & \multicolumn{1}{c|}{\cellcolor[HTML]{CBD8FF}\textbf{41.3}} & \cellcolor[HTML]{CBD8FF}\textbf{39.8} & \multicolumn{1}{c|}{\cellcolor[HTML]{CBD8FF}41.8} & \cellcolor[HTML]{CBD8FF}40.7 \\ \bottomrule
\end{tabular}
\label{tab:all_ab}
\end{table}

\section{Generalization Study of Quantization model}
In our experiments, in order to adapt SAM to specific downstream tasks, calibration data was sampled from the corresponding training dataset, and prompt design is also task-dependent. 
For example, in instance segmentation task, we use COCO datasets with boxes from detectors as prompt, while in semantic segmentation task, we use ADE20K dataset with valid positive grid points as prompt.
To investigate whether our method can maintain the generalization ability of quantized SAM, we performed experiments using quantized SAM on semantic segmentation for instance segmentation.
As shown in Table \ref{tab:generalization}, models learned in instance segmentation on COCO also excel in semantic segmentation tasks on ADE20K, demonstrating strong generalization beyond its learning prompts and datasets.

\begin{table}[ht]\scriptsize
\centering
\setlength{\tabcolsep}{5pt}
\caption{
Generalization study of our methods.
}
\begin{tabular}{@{}c|c|c|c@{}}
\toprule
\begin{tabular}[c]{@{}c@{}}Test\\ (Dataset/Prompt)\end{tabular} & \begin{tabular}[c]{@{}c@{}}Calibration\\ (Dataset/Prompt)\end{tabular} & \begin{tabular}[c]{@{}c@{}}W6A6\\ (Baseline: 32.65)\end{tabular} & \begin{tabular}[c]{@{}c@{}}W4A4\\ (Baseline: 31.85)\end{tabular} \\ \midrule
 & ADE20K/Points & 33.04($\uparrow$ 0.39) & 32.53($\uparrow$ 0.68) \\ \cmidrule(l){2-4} 
\multirow{-2}{*}{ADE20K/Points} & \cellcolor[HTML]{EFEFEF}COCO/Bboxes & \cellcolor[HTML]{EFEFEF}32.99(\textcolor{green}{$\uparrow$ 0.34}) & \cellcolor[HTML]{EFEFEF}32.15(\textcolor{green}{$\uparrow$ 0.30}) \\ \bottomrule
\end{tabular}
\label{tab:generalization}
\end{table}

\end{document}